\documentclass[10pt,journal,compsoc]{IEEEtran}
\ifCLASSOPTIONcompsoc
  \usepackage[nocompress]{cite}
  \usepackage[caption=false,font=footnotesize,labelfont=sf,textfont=sf]{subfig}
\else
  \usepackage{cite}
   \usepackage[caption=false,font=footnotesize]{subfig}
\fi
\usepackage{amsmath}
\usepackage{amssymb}
\usepackage{algorithmic}
\usepackage{array}
\usepackage{dblfloatfix}
\ifCLASSOPTIONcaptionsoff
  \usepackage[nomarkers]{endfloat}
 \let\MYoriglatexcaption\caption
 \renewcommand{\caption}[2][\relax]{\MYoriglatexcaption[#2]{#2}}
\fi
\usepackage{url}

\usepackage{wasysym}
\usepackage{multirow}
\usepackage{floatrow}
\usepackage{makecell,pict2e}
\usepackage[ruled]{algorithm}
\usepackage{balance}
\newsavebox{\tempsavebox}

\ifCLASSINFOpdf
  \usepackage[pdftex]{graphicx}
  \graphicspath{{./figures/pdf/}{./figures/jpeg/}}
  \DeclareGraphicsExtensions{.pdf,.jpeg,.png,.jpg}
\else
  \usepackage[dvips]{graphicx}
  \graphicspath{{./figures/eps/}}
  \DeclareGraphicsExtensions{.eps}
\fi

\usepackage{xcolor}

\begin{document}
\title{Boosting Occluded Image Classification via Subspace Decomposition Based Estimation of Deep Features}

\author{Feng~Cen,
        and Guanghui~Wang, \IEEEmembership{Senior Member, ~IEEE}
	\thanks{F. Cen is with the Department of Control Science \& Engineering,
		College of Electronics and Information Engineering,
		Tongji University, Shanghai 201804, China
		Email: feng.cen@tongji.edu.cn}
	\thanks{G. Wang is with the Department of Electrical Engineering and Computer Science,
		University of Kansas, Lawrence, KS 66045.  Email: ghwang@ku.edu}
	\thanks{Manuscript received xxxx, 2018; revised xxxx, 2018.}}

\IEEEtitleabstractindextext{%
\begin{abstract}
Classification of partially occluded images is a highly challenging computer vision problem even for the cutting edge deep learning technologies.
To achieve a robust image classification for occluded images, this paper proposes a novel scheme using subspace decomposition based estimation (SDBE).
The proposed SDBE-based classification scheme first employs a base convolutional neural network to extract the deep feature vector (DFV) and then utilizes the SDBE to compute the DFV of the original occlusion-free image for classification.
The SDBE is performed by projecting the DFV of the occluded image onto the linear span of a class dictionary (CD) along the linear span of an occlusion error dictionary (OED).
The CD and OED are constructed respectively by concatenating the DFVs of a training set and the occlusion error vectors of an extra set of image pairs.
Two implementations of the SDBE are studied in this paper: the $l_1$-norm and the squared $l_2$-norm regularized least-squares estimates.
By employing the ResNet-152, pre-trained on the ILSVRC2012 training set, as the base network, the proposed SBDE-based classification scheme is extensively evaluated on the Caltech-101 and ILSVRC2012 datasets.
Extensive experimental results demonstrate that the proposed SDBE-based scheme dramatically boosts the classification accuracy for occluded images, and achieves around $22.25\%$ increase in classification accuracy under $20\%$ occlusion on the ILSVRC2012 dataset.
\end{abstract}

}

\maketitle

\IEEEpeerreviewmaketitle

\IEEEraisesectionheading{\section{Introduction}\label{sec:introduction}}
\IEEEPARstart{O}{cclusion} occurs in many real world images.
For human vision system, recognizing partially occluded objects is not a tough mission.
In computer vision domain, however, this is still a highly challenging task even for deep convolutional neural networks (CNNs) which have achieved huge success in many computer vision tasks recently~\cite{krizhevsky2012imagenet, he2018learning, zhang2018bpgrad, wei2017cross, han2017cnns, xu2019adversarially}.
The state-of-the-art CNNs usually involve over tens of millions of parameters \cite{szegedy2015going, he2016deep, simonyan2014very, ma2018mdcn} such that a vast amount of data are required in training even for the classification of occlusion-free images.
For the classification of occluded images, the training dataset has to be enlarged multiple times to cover the variations caused by occlusion.
To collect a very large number of occluded images is, however,  difficult in real applications.
A popular choice is to train the networks directly on an occlusion-free dataset or a dataset containing few occluded images. 
Unfortunately, the deep features generated by most CNN networks are sensitive to occlusion. 
As a consequence, little benefit can be gained to the classification of occluded images.

\begin{figure}
	\centering
	\fontsize{8}{8}\selectfont
	\centering
	\includegraphics[width=0.95\linewidth]{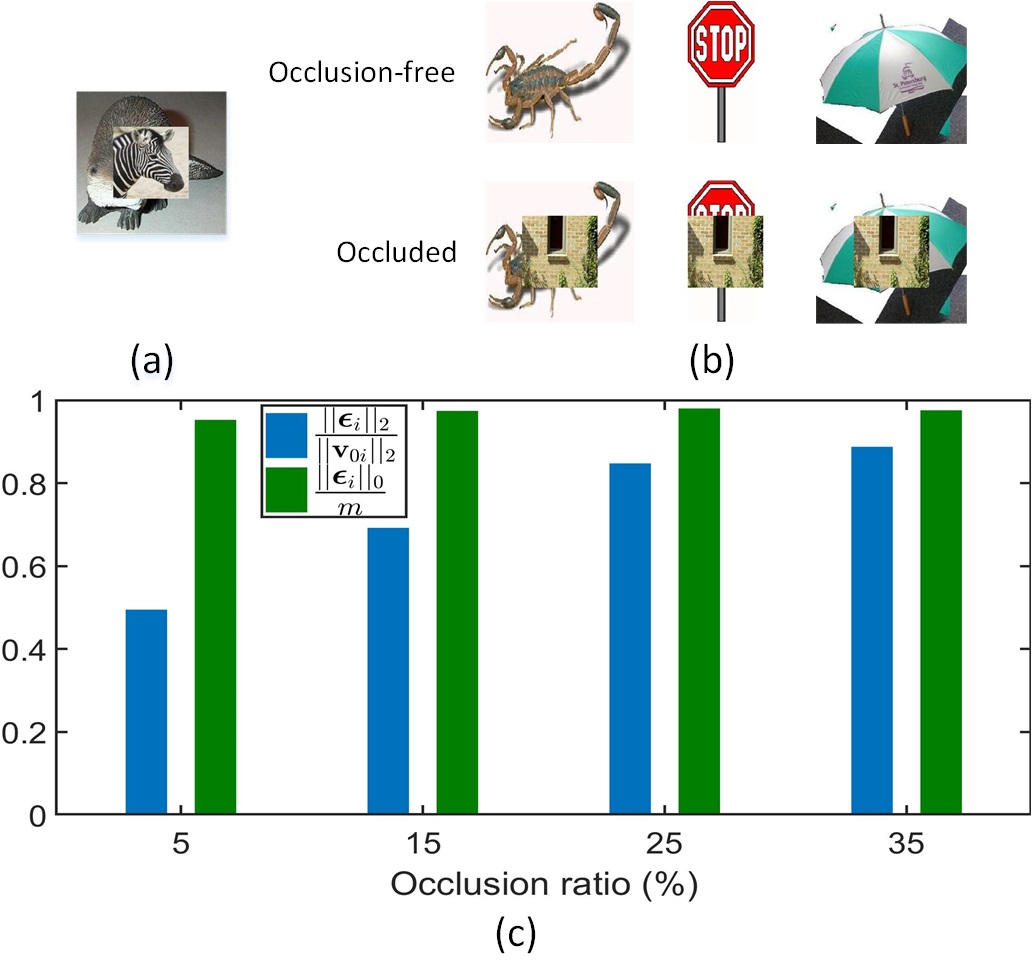}
	\caption{(a) Example of an occluded query image. (b) Examples of extra image pairs.  
		(c)Illustration of $||\boldsymbol{\epsilon}_i||_2$ and $||\boldsymbol{\epsilon}_i||_0$ of an occluded image with respect to occlusion ratio. The DFVs are extracted by using the ResNet-152 network \cite{he2016deep} and then normalized to have unit $l_2$-norm. To make a fair comparison, $||\boldsymbol{\epsilon}_i||_2$ and $||\boldsymbol{\epsilon}_i||_0$ are normalized with respect to $||\mathbf{v}_{0i}||_2$ and the number of the DFV entries $m$, respectively.}
	\label{fig:extraimg_errorillustration}
\end{figure}

In a typical application of classification, the training dataset usually contains much fewer occluded images than occlusion-free images, and sometimes, one may have some occluded images irrelevant to the task-specified categories.  
To address the classification of occluded images for this kind of scenario, this paper addresses the problem of classifying partially occluded images by exploiting the help of a set of extra image pairs.
Each extra image pair includes an occlusion-free image and an occluded image.
The occluded image is contaminated with an occlusion pattern involved in the occluded query images.
Under such a setting, the extra image pairs can provide auxiliary information about the occlusions of the occluded query images.
Examples of an occluded query image and extra image pairs are shown in \figurename{~\ref{fig:extraimg_errorillustration}} (a) and (b), respectively.

The research is originally motivated by cloud-based deep learning applications \cite{hosseini2016cloud,Deng2013Recent,Polishetty2017A, cen2019dictionary}.
With the development of deep neural network accelerator chips \cite{isscc_2016_chen_eyeriss}, the CNN-based feature extraction, used to be conducted at the cloud or server end due to its high computational complexity, are gradually shifting to the terminal or mobile end.
To save the bandwidth of transmission in such a situation, only the deep features, which have less bandwidth consumption and safer in terms of protection of private information, rather than the images will be sent to the cloud for further processing, such as classification, identification, and object detection.
Accordingly, we have to tackle the occlusion in the deep feature space at the cloud end.
Therefore, in this paper, instead of removing occlusion in the image space, we focus on alleviating the negative impact of occlusion on classification in the deep feature space.

Let $\mathbf{v}_{0i}$ be the deep feature vector (DFV) of the $i$th occlusion-free image $\mathbf{y}_{0i}$ and $\mathbf{v}_i$ the DFV of the occluded image $\mathbf{y}_i$ 
acquired by contaminating the image $\mathbf{y}_{0i}$ with a contiguous patch $\mathbf{z}$.
Then, we have 
\begin{equation}\label{equ:dpvec_dcmp}
\mathbf{v}_i=\mathbf{v}_{0i}+\boldsymbol{\epsilon}_i,
\end{equation}
where $\boldsymbol{\epsilon}_i$ denotes the occlusion error vector (OEV) reflecting the variation caused by the occlusion.
A simple and intuitive solution to recognize the occluded image with a classifier trained on occlusion-free images is first estimating $\mathbf{v}_{0i}$ from $\mathbf{v}_i$ and then feeding $\hat{\mathbf{v}}_{0i}$, the estimation of $\mathbf{v}_{0i}$, into the classifier to determine the category.

Recovering $\mathbf{v}_{0i}$ from $\mathbf{v}_i$ is, however, a challenge work. 
Even though many efforts have been made to understand the deep representations \cite{bau2017network,zeiler2014visualizing,mahendran2015understanding,alain2016understanding}, the correspondence between the occlusion and the changes in the DFV is still far from being clear.
The CNNs is actually a nonlinear and holistic transformation from the image space to the deep feature space.
Any small local variations in the image space can cause large holistic changes in the deep feature space.
As can be seen in \figurename{~ \ref{fig:extraimg_errorillustration}}(c), even for $5\%$ occlusion, a small amount of occlusion, around $95\%$ entries of $\boldsymbol{\epsilon}_i$ are nonzero and the $l_2$-norm of $\boldsymbol{\epsilon}_i$ is close to $50\%$ $l_2$-norm of $\mathbf{v}_{0i}$, a very high level of relative energy.

In this paper, we observe that in the deep feature space, $\boldsymbol{\epsilon}_i$ is a structured error clustering at a place outside the \emph{class subspace}, the linear span of the DFVs of occlusion-free images.
This indicates that $\boldsymbol{\epsilon}_i$ lies in a low-dimensional subspace, named \emph{occlusion error subspace}, nearly independent of the \emph{class subspace}.
Inspired by this observation, we propose a subspace decomposition based estimation (SDBE) to extract $\mathbf{v}_{0i}$ by finding a constrained projection of $\mathbf{v}_i$ onto the \emph{class subspace} along the \emph{occlusion error subspace}.
In practice, we use the linear span of the DFVs of a training set, named class dictionary (CD), to approximate the \emph{class subspace} and the linear span of the OEVs of the extra image pairs, named occlusion error dictionary (OED), to roughly represent the \emph{occlusion error subspace}.
The $l_1$-norm and the squared $l_2$-norm regularizations are studied to constrain the projection and least-squares (LS) optimization is employed to compute the constrained projection.

Based on the proposed SDBE approaches, a classification scheme is developed in this paper.
In the proposed SDBE-based classification scheme, a base CNN is employed to map the image to a deep feature space linearly separable for the occlusion-free images, and then, the proposed SDBE approaches are applied to project the DFV onto the \emph{class subspace}.
The base CNN can be trained on publicly available large-scale datasets or task-specified datasets and the SDBE is applicable to both occluded images and occlusion-free images (corresponding to zero occlusion).
Therefore, the proposed SDBE-based classification scheme is a unified scheme for the classification of both occluded images and occlusion-free images over any image datasets.

Our main contributions are summarized as follows.
\begin{enumerate}
	\item We observe that the DFVs and OEVs lie in distinct low-dimensional subspaces. This observation provides a useful clue to fully understanding the representation learned by the CNNs.
	\item We propose a novel SDBE to compute the DFV of the the original occlusion-free image and introduce two implementations for the SDBE: the $l_1$-norm LS minimization and the squared $l_2$-norm LS minimization.
	\item Based on the proposed SDBE approaches, we propose an SDBE-based classification scheme and present extensive experiments on publicly available small-scale and large-scale datasets.
	The experimental results demonstrate significant improvement over the state-of-the-art conventional CNN-based schemes. To the best of our knowledge, this is the first study exhibiting impressive classification results for occluded images on a general purpose large-scale image dataset.
	\item Although end-to-end learning is pursued in many research works, the successful integration of the CNNs and the classical learning approaches provides a new perspective to other similar problems. 
	We also introduce an implementation in linear network layer form for the squared $l_2$-norm based SDBE approach to facilitate the adaption of the proposed scheme to many pervasive implementation frameworks of the CNNs.
\end{enumerate}

The rest of this paper is organized as follows.
Section \ref{sec:Relatedwork} briefly reviews the related works. 
Section \ref{sec:Proposedscheme} describes the proposed SDBE approaches and SDBE-based classification scheme in detail.
Section \ref{sec:Exps} presents the experimental results.
Section \ref{sec:further_discussion} compares two implementations of the proposed SDBE in detail.
Finally the paper is concluded in Section \ref{sec:Conclusion}.

\section{Related works}\label{sec:Relatedwork}
The work in this paper is partially related to the works in the following fields.
\subsection{Signal recovery}
In signal processing community, compressed sensing is an extensively studied signal recovery approach.
However, it is unable to apply the compressed sensing straightforwardly to estimate $\mathbf{v}_{0i}$.
In most compressed sensing research works, the $\boldsymbol{\epsilon}_i$ is assumed to be a random noise term.
For random noise, the existing theory \cite{donoho2006stable,nguyen2013robust,ben2010cramer,ben2010coherence} suggests that the $\boldsymbol{\epsilon}_i$ needs to be either sparse or small, which is in contradiction to the observation in \figurename{~\ref{fig:extraimg_errorillustration}}(c).
 
The closely related theoretical works in compressed sensing include \cite{studer2012recovery,studer2014stable}, where the noise term is assumed to be structured noise.
These works, however, only focused on signal recovery and pursued a sparsest solution.
Unlike the signal recovery, the goal of classification is not an exact recovery of original signal but rather an estimation leading to correct classification.
This target implies that an estimation even with pretty large deviation from the original signal can be acceptable for the classification.

\subsection{Sparse representation-based classification}
The most popular approach robustifying the occluded image classification in computer vision is sparse representation-based classification, proposed by Wright \textit{et al.} \cite{wright2009robust}.
In their work, the occluded face image was first coded via $l_{1}$-norm minimization as a sparse linear combination of the expanded dictionary.
Then, the classification is conducted by searching for which class of training samples could result in a minimum reconstruction error with the sparse coding coefficients.
Following Wright's work, many researchers worked towards improving the sparse representation-based classification accuracy under various conditions \cite{deng2012extended,ou2014robust,zhang2015mixed,zhao2016corrupted} .

The weaknesses of these works lie in that they are focused on the image space or the linear transformation of the image space and are only shown to be effective in face recognition, a narrow subfield of image classification, over small-scale datasets.

\subsection{Generative model}
With the rapid development of deep learning, a lot of effort has been made to apply deep generative model to cope with partially occluded or partially missing image recently.
In~\cite{pathak2016context,yeh2016semantic,nguyen2017plug,yang2017high,yu2018generative}, deep dilated convolution networks were used to yield the missing portion of an image based on the framework of generative adversarial networks (GANs) and combined with a loss function related to the reconstruction error.
In~\cite{bao2017cvae}, variational auto-encoder and GAN were combined to generate the missing portion of an image.
In~\cite{van2016pixel}, pixel recurrent neural networks were proposed for image completion.
These approaches, however, need to know the shape and position of the missing portion in advance.
In~\cite{tang2010gated,Tang2012Robust}, restricted Boltzman machine based models were used to learn the structure of the occluders.
In~\cite{xie2012image,cheng2015robust}, denoising auto-encoder based models were exploited to map a corrupted image to a corruption-free image.
In~\cite{zhao2018robust}, a robust LSTM-autoencoders model is combined with GAN to produce the occluded portion of the face image for face recognition. 

The weaknesses of these works lie in the following aspects.
First, these works only yielded improved results in face recognition and did not exhibit promising results for general occluded images. 
Second, these works are not suited to future cloud-based applications since they attempt to restore images in the image space.
Third, these works usually require a large number of partially occluded or missing images and a time-consuming training procedure to train the generative model.
Finally, for new occlusion patterns, these works require a re-training or fine-tuning, usually complex and time-consuming, of the generative model.
In contrast, the proposed SDBE-based classification scheme handles the occlusion in the deep feature space and  requires much fewer occluded images for training. It can be easily adapted to new occlusion patterns, and exhibits superior classification results for general occluded images. 

\subsection{Deep feature manipulation}
Understanding the relationship between the changes in the image space and the consequences in the deep feature space is still a challenge.
Some research effort has been made to manipulate the deep features to tackle the variation in the image space~\cite{wen2016latent, upchurch2017deep,chen2017stylebank,chen2018facelet, gao2016novel}, recently.
Wen \textit{et al.}~\cite{wen2016latent} introduced a latent factor fully connected (LF-FC) layer (a linear transformation matrix) to extract the age-invariant deep features from convolutional features for aging face recognition.
Li \textit{et al.}~\cite{li2017perceptual} proposed to use the GAN in deep feature space to generate super-resolved representation for small object detection.
Chen \textit{et al.}~\cite{chen2017stylebank} fed the intermediate deep feature into multiple convolution filter banks to perform image style transfer.
Chen \textit{et al.}~\cite{chen2018facelet} employ multiple fully convolutional layers to manipulate the middle-level convolutional representations for face portrait transfer.

Although these works did not directly relate to the classification of occluded images, they showed a trend to cope with the variation of an image in the deep feature space.
In these works, however, the transformation or mapping of the deep feature is achieved by using neural network layers, which are trained with back-propagation algorithms, thereby requiring a large number of images in training.
In contrast, the proposed SDBE approaches are based on a theoretical basis and adopt a simpler learning approach, requiring much fewer training images.

\section{Proposed scheme}\label{sec:Proposedscheme}
\begin{figure*}
	\centering
	\fontsize{6}{8}\selectfont
	\includegraphics[width=0.65\linewidth]{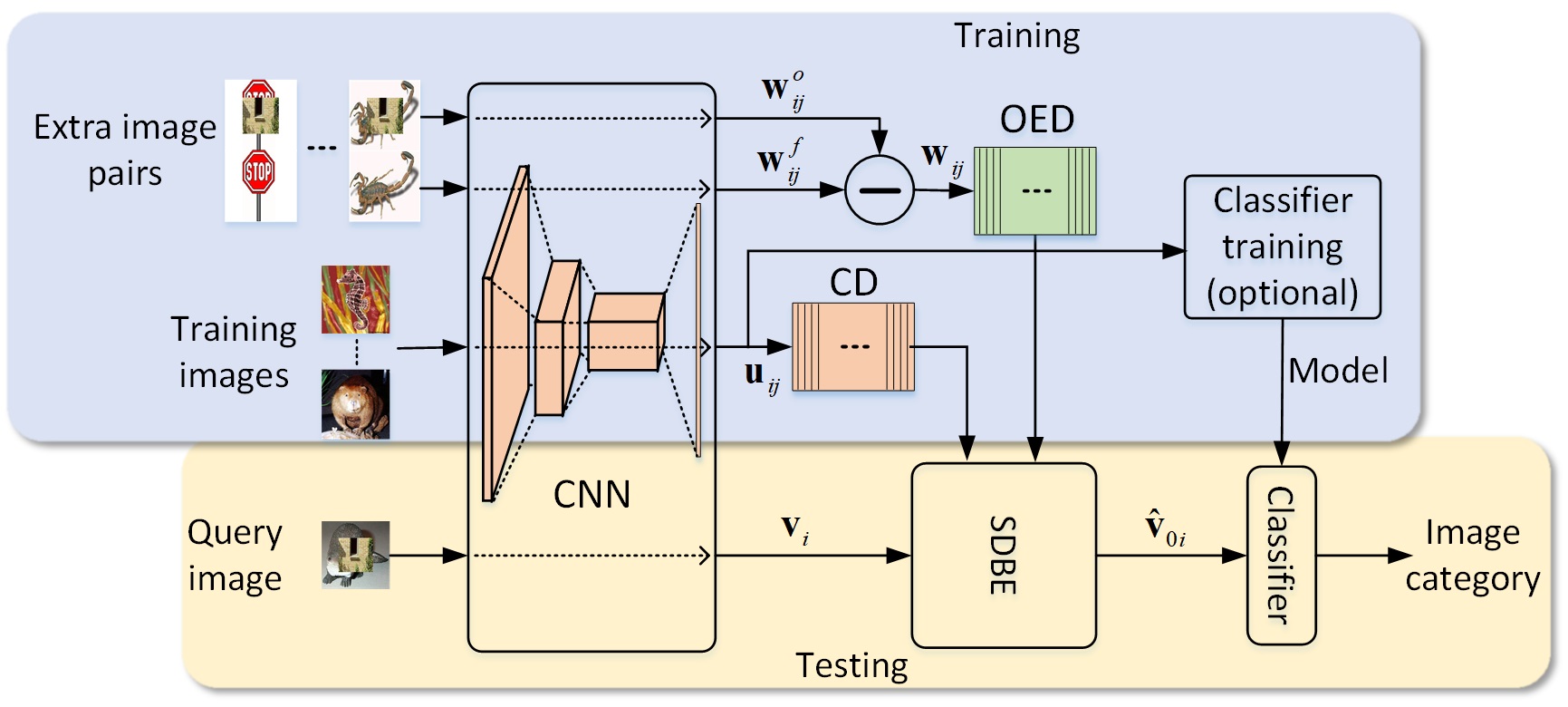}
	\caption{Flowchart of the proposed SDBE-based classification scheme. 
		The proposed SDBE is inserted into the processing chain of classification between the base CNN and the classifier as a post-processing procedure for the DFVs.
	}
	\label{fig:clsfct_scheme}	
\end{figure*}
The proposed SDBE-based classification scheme is shown in \figurename{~\ref{fig:clsfct_scheme}}.
A base CNN (excluding the prob layer and the last fully connected linear layer of the original CNN network) is employed to extract the DFVs.
In the training phase, a CD and an OED are constructed by using the DFVs extracted from the training images and extra image pairs, respectively.
In the testing phase, the SDBE, as a post-processing procedure for the DFV, is employed to mitigate the error, induced by occlusion, of the DFV $\mathbf{v}_i$ with the help of the CD and OED.
Following the SDBE, a classifier will determine the category of the DFV $\hat{\mathbf{v}}_{0i}$ estimated by the SDBE.

The classifier training procedure is optional since the classifier can be any conventional classifiers, such as support vector machine (SVM), softmax, and nearest neighbor (NN) \cite{duda2001pattern}, trained by using either the column vectors of the CD or other DFVs extracted from a task-specified image dataset, or the original softmax classifier of the base CNN, which is actually trained along with the base CNN.

The base CNN can be trained on large-scale publicly available datasets or a task-specified dataset for a better separability of the deep feature space. 
It can be constructed with any modern network structures, such as ResNet~\cite{he2016deep}, GoogLeNet~\cite{szegedy2015going}, and VGG~\cite{simonyan2014very}, as long as it is well trained for classification. 
The details of construction and training of the base CNNs are out of the scope of this paper.
Let the size of input image and the number of the entries of the output DFV be $h\times w$ and $m$, respectively.
The base CNNs perform a nonlinear mapping from three-channel color image space to the deep feature space, $f:\mathbb{R}^{h\times w\times 3}\longmapsto \mathbb{R}^{m}$.
Then, we have $\mathbf{v}_i=f(\mathbf{y}_i)$ and $\mathbf{v}_{0i}=f(\mathbf{y}_{0i})$.
We should note that the OEV $\boldsymbol{\epsilon}_i$ is not a mapping of the occlusion patch $\mathbf{z}$, i.e., $\boldsymbol{\epsilon}_i\neq f(\mathbf{z})$\footnote{$\boldsymbol{\epsilon}_i$ is associated with not only the physical occlusion errors in the image but also the image, since the deep feature is a kind of nonlinear holistic feature.}.

\subsection{SDBE}\label{sec:SDBE}
Suppose that we have a training set with $n_{\mathbf{A}}=\sum_{i=1}^{K_{\mathbf{A}}}n_i$ images collected from $K_{\mathbf{A}}$ categories and a set of extra image pairs with $p_{\mathbf{B}}=\sum_{i=1}^{K_{\mathbf{B}}}p_i$ image pairs associated with $K_{\mathbf{B}}$ occlusion patterns.
Here, $n_{i}$ is the number of training images belonging to the $i$th category and $p_{i}$ the number of extra image pairs associated with the $i$th occlusion pattern. 
The occlusion pattern is defined as the occlusions with the same texture, shape, size, and location on the image.
Let $\mathbf{u}_{ij}\in\mathbb{R}^{m}, j=1,2,\cdots,n_{i}$ be the DFV of the $j$th training image in the $i$th category.
The extra image pairs each consists of an occlusion-free image and an occluded image.
Let $\mathbf{w}_{ij}^f\in\mathbb{R}^{m}$ and $\mathbf{w}_{ij}^o\in\mathbb{R}^{m}, j=1,2,\cdots,p_{i}$ be the DFVs of the occlusion-free image and occluded image of the $j$th extra image pair associated with the $i$th occlusion pattern, respectively. 
Then, the OEV between $\mathbf{w}_{ij}^o$ and  $\mathbf{w}_{ij}^f$ is given by
\begin{equation}\label{equ:wij_dfn}
\mathbf{w}_{ij}=\mathbf{w}_{ij}^o-\mathbf{w}_{ij}^f\;.
\end{equation} 

\subsubsection{Class subspace vs. occlusion error subspace}
It is easy to observe that, extracted from a well trained base CNN, the DFVs of the occlusion-free images in each category usually locate in a compact low-dimensional cluster, called class cluster.
Let $\mathcal{A}_i$ denote the linear span of the $i$th class cluster and $\mathcal{A}=\sum_i \mathcal{A}_i$ the sum over all the categories used in a classification task.
Then, $\mathcal{A}$ can be regarded as a low-dimensional subspace, called \emph{class subspace}
\footnote{The \emph{class subspace} can be determined by using the principal component analysis (PCA), as shown in \figurename{~\ref{fig:relationship_A_B}\subref{fig:pca_A_B}}}.
In addition, we assume that the OEVs incurred by the same occlusion pattern fall into a low-dimensional subspace.
Let $\mathcal{B}_i$ denote the subspace associated with the $i$th occlusion pattern and $\mathcal{B}=\sum_i \mathcal{B}_i$ the sum of all the subspaces associated with the occlusion patterns involved in the set of extra image pairs.
Obviously, the $\mathcal{B}$ spans a subspace, named \emph{occlusion error subspace}.
For simplicity, hereinafter we reuse, without ambiguity in context, the notation of a subspace to denote the basis of that subspace, e.g., the $\mathcal{A}$ can also stand for the basis of the subspace $\mathcal{A}$.

\subsubsection{Decomposition over class subspace and occlusion error subspace}
Suppose that $\mathcal{A}$ is linearly independent of $\mathcal{B}$ and the DFV $\mathbf{v}_i$ of the query image lies in the subspace $\mathcal{V}=\mathcal{A}\oplus\mathcal{B}$, where $\oplus$ denotes direct sum.
Then, $\mathbf{v}_i$ has a unique decomposition with the form of \cite[Theorem 1.5]{roman2007advanced} 
\begin{equation}\label{equ:subspace_dcmpstn}
\mathbf{v}_i=\mathcal{A}\boldsymbol{\alpha}+\mathcal{B}\boldsymbol{\beta},
\end{equation}
where $\boldsymbol{\alpha}$ and $\boldsymbol{\beta}$ are decomposition coefficient vectors.
The class part $\mathcal{A}\boldsymbol{\alpha}$, a projection of $\mathbf{v}_i$ onto $\mathcal{A}$ along $\mathcal{B}$, is equal to $\mathbf{v}_{0i}$.

Model (\ref{equ:subspace_dcmpstn}) is, however, almost practically unrealizable. 
The reasons are as below: 
first, figuring out the exact spans or bases of the \emph{class subspace} and \emph{occlusion error subspace} is unattainable in real applications;
second, for real image data, $\mathcal{A}$ and $\mathcal{B}$ are not exactly independent.

In this paper, instead of finding the exact spans or bases, we utilize the linear span of $\mathbf{A}_i=[\mathbf{u}_{i1},\mathbf{u}_{i2},...,\mathbf{u}_{in_{i}}]\in\mathbb{R}^{m\times n_i}$ and the linear span of  $\mathbf{B}_i=[\mathbf{w}_{i1},\mathbf{w}_{i2},...,\mathbf{w}_{ip_{i}}]\in\mathbb{R}^{m\times p_i}$
to approximate $\mathcal{A}_i$ and $\mathcal{B}_i$, respectively.
Apparently, with such an approximation, error will be introduced into the decomposition.
To explicitly account for the error, we add a noise term $\mathbf{n}$ to model (\ref{equ:subspace_dcmpstn}).
Then, we have
\begin{equation}\label{equ:approx_subspace_dcmpstn}
\mathbf{v}_i=\mathbf{A}\boldsymbol{\alpha}+\mathbf{B}\boldsymbol{\beta}+\mathbf{n}
\end{equation}
with $\boldsymbol{\alpha}=[\boldsymbol{\alpha}_1^T,...,\boldsymbol{\alpha}_i^T,...,\boldsymbol{\alpha}_{K_{\mathbf{A}}}^T]^T$ and $\boldsymbol{\beta}=[\boldsymbol{\beta}_1^T,..., \boldsymbol{\beta}_i^T,...,\boldsymbol{\beta}_{K_{\mathbf{B}}}^T]^T$, where $\mathbf{A}=[\mathbf{A}_1,...,\mathbf{A}_i,...,\mathbf{A}_{K_{\mathbf{A}}}]$ and $\mathbf{B}=[\mathbf{B}_1,..., \mathbf{B}_i, ...,\mathbf{B}_{K_{\mathbf{B}}}]$ are the CD and OED, respectively, and $\boldsymbol{\alpha}_i$ and $\boldsymbol{\beta}_i$ are the vectors of decomposition coefficients related to $\mathbf{A}_i$ and $\mathbf{B}_i$, respectively.
Then, $\mathbf{v}_{0i}$ can be estimated by
\begin{equation}\label{equ:dfv_rcvry}
\hat{\mathbf{v}}_{0i}=\mathbf{A}\boldsymbol{\alpha}.
\end{equation}

In a more concise form, equation (\ref{equ:approx_subspace_dcmpstn}) can be written  as
\begin{equation} \label{equ:prblm_frmltn_rewriting}
\mathbf{v}_i=\mathbf{D}\boldsymbol{\omega}+\mathbf{n}
\end{equation}
with the concatenated dictionary $\mathbf{D}=[\mathbf{A}\; \mathbf{B}]$ and the stacked vector $\boldsymbol{\omega}_i=[\boldsymbol{\alpha}^T\; \boldsymbol{\beta}^T]^T$.

\begin{figure}
	\centering
	\fontsize{8}{8}\selectfont
	\subfloat[]{
		\centering
		\includegraphics[width=0.8\linewidth]{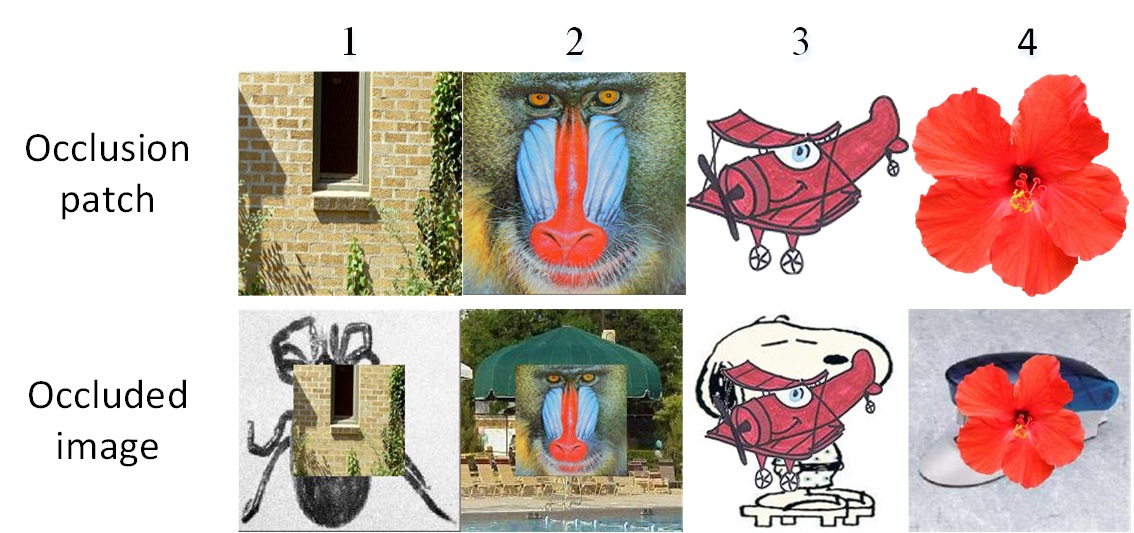}
	}\\
	\subfloat[]{\label{fig:pca_A_B}
		\includegraphics[width=0.5\linewidth]{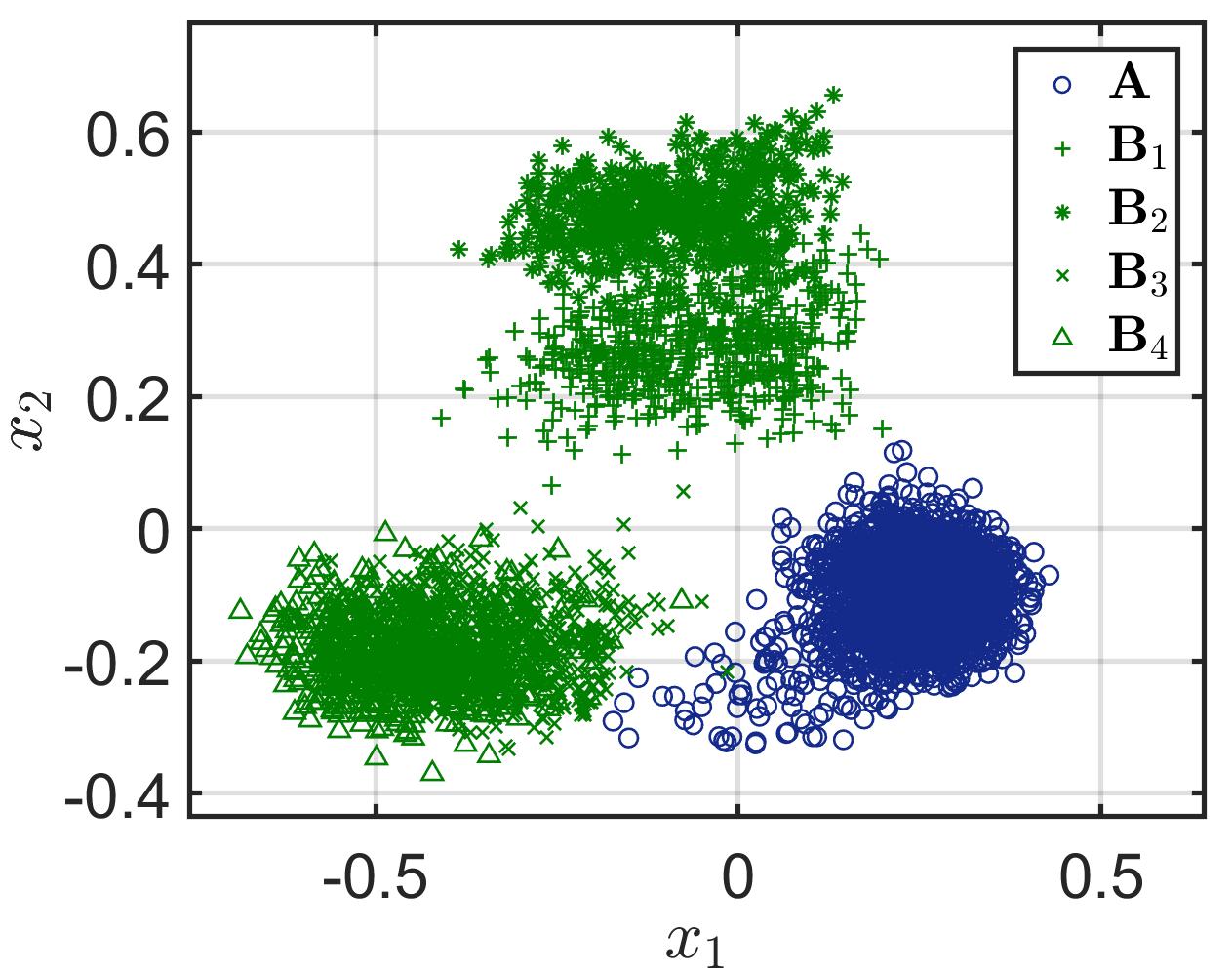}
	}
	\subfloat[]{\label{fig:correlation_A_B_AB}
		\includegraphics[width=0.5\linewidth]{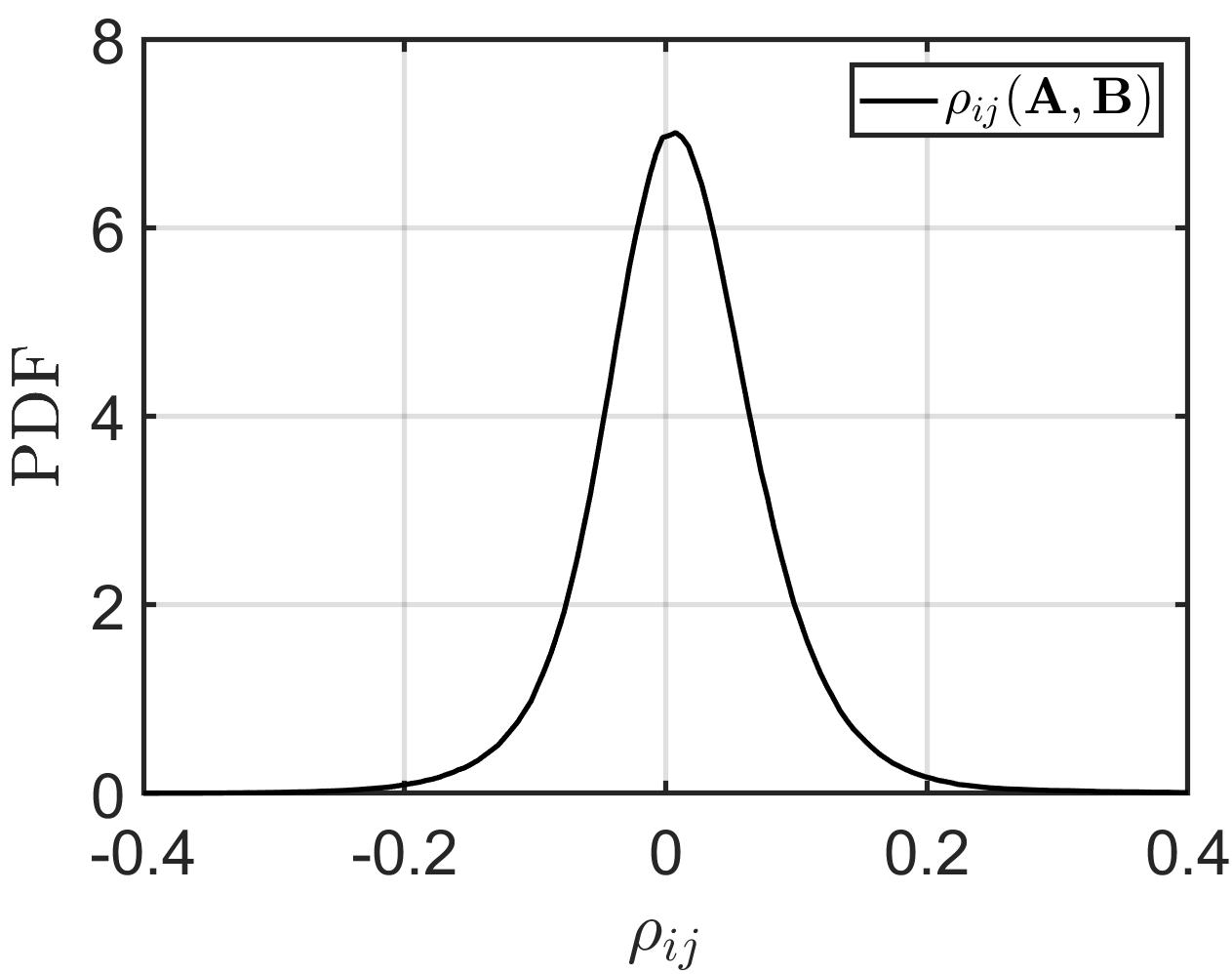}
	}
	\caption{Illustration of linear independence between \emph{class subspace} and \emph{occlusion error subspace}. (a) Four occlusion patches used to construct $\mathbf{B}$ and examples of occluded images.
		(b) Projection of column vectors of $\mathbf{A}$ and $\mathbf{B}_k$'s onto the first two largest principle components of $\mathbf{D}$, $x_1$ and $x_2$.
		The CD $\mathbf{A}$ comes from the experiment of $25\%$ occlusion in Section \ref{exp:hyprprmtr} and OED $\mathbf{B}$ is construct by using the occlusion patches in (a) with the same setting as the OED in the experiment of $25\%$ occlusion in Section \ref{exp:hyprprmtr} .
		$\mathbf{B}_k$ is associated with the $k$th occlusion patch in (a). 
		(c) Probability density function (PDF) of $\rho_{ij}(\mathbf{A},\mathbf{B})$.
	}
	\label{fig:relationship_A_B} 
\end{figure}

In \figurename{~\ref{fig:relationship_A_B}}, we use a toy example to explain the preceding description on the \emph{class subspace} and \emph{occlusion error subspace}.
From \figurename{~\ref{fig:relationship_A_B}\subref{fig:pca_A_B}}, we can easily observe that the column vectors of $\mathbf{A}$ cluster together (blue cluster) and the column vectors of each $\mathbf{B}_i$ forms a cluster (green clusters).
This observation indicates that the subspace $\mathbf{A}$ is distinct from the subspace $\mathbf{B}$ and $\mathbf{B}$ has fine-grained low-dimensional structures.

In \figurename{~\ref{fig:relationship_A_B}\subref{fig:correlation_A_B_AB}}, due to lack of proper method to measure the degree of linear independence, we adopt the correlation  between two vectors to approximately indicate the degree of linear independence on account of the equivalence between uncorrelation and independence for two vectors~\cite{rodgers1984linearly}.
The correlation is measured with the Pearson correlation coefficient.
Let $\mathbf{X}\in \mathbb{R}^{m\times n_x}$ and $\mathbf{Y}\in \mathbb{R}^{m\times n_y}$ be two matrices and $\mathbf{x}_i=[\cdots,x_{li},\cdots]^T$ and $\mathbf{y}_i=[\cdots,y_{li},\cdots]^T$ be the $i$th column vectors of $\mathbf{X}$ and $\mathbf{Y}$, respectively.
The Pearson correlation coefficient $\rho_{ij}(\mathbf{X},\mathbf{Y})$ between $\mathbf{x}_i$ and $\mathbf{y}_i$ can be written as
\begin{equation}\label{equ:correlation_def}
\rho_{ij}(\mathbf{X},\mathbf{Y})=\frac{\sum_{l=1}^{m}(x_{li}-\bar{x}_i)(y_{lj}-\bar{y}_j)}{\sqrt{\sum_{l=1}^{m}(x_{li}-\bar{x}_i)^2}\sqrt{\sum_{l=1}^{m}(y_{lj}-\bar{y}_j)^2}},
\end{equation}
where $\bar{x}_i=\frac{1}{m}\sum_{l=1}^{m}x_{li}$ and $\bar{y}_j=\frac{1}{m}\sum_{l=1}^{m}y_{lj}$ are the sample means of $\mathbf{x}_i$ and $\mathbf{y}_j$, respectively. 
A high magnitude of $\rho_{ij}(\mathbf{X},\mathbf{Y})$ indicates a strong correlation between $\mathbf{x}_i$ and $\mathbf{y}_i$.

As can be seen in \figurename{~\ref{fig:relationship_A_B}\subref{fig:correlation_A_B_AB}}, $\rho_{ij}(\mathbf{A},\mathbf{B})$ locates around zero with a mean magnitude of $0.0514$.
This indicates that the $\mathbf{A}$ and $\mathbf{B}$ are close to be uncorrelated and independent.

These results substantiate our assumption that the DFVs and OEVs locate in distinct and linearly independent low-dimensional subspaces. 

Although the assumption of independence does not hold strictly in practice, close to independence can occur with high probability, owing to the different value ranges for the DFV elements and OEV elements.
The value range for the DFV elements is asymmetric, whereas that for the OEV elements is symmetric.
For instance, the ReLU layer has a non-negative range of output, as a result, the output DFV of the ReLU layer only has non-negative elements.
On the contrary, the elements of the OEV can take both positive and negative values since no constraint is set on the range of the values.

A popular method to solve equation (\ref{equ:prblm_frmltn_rewriting}) is the LS estimates~\cite{schmidt2005least}.
Generally, equation (\ref{equ:prblm_frmltn_rewriting}) has multiple solutions \footnote{In typical applications, the column rank of $\mathbf{D}$ is smaller than $n_\mathbf{A}+p_\mathbf{B}$}, not all of which can improve the performance.
Nevertheless, by imposing proper restriction on the decomposition coefficients, a solution with the class part $\mathbf{A}\boldsymbol{\alpha}$ close to $\mathbf{v}_{0i}$ and falling into the correct class cluster can be achieved, since $\mathbf{A}$ and $\mathbf{B}$ closely satisfy the preceding assumption on subspace and linear independence.
Regularization is a common approach to constrain the coefficients.
With the regularization, the solution of equation (\ref{equ:prblm_frmltn_rewriting}) can be written in a general form as
\begin{equation}\label{equ:general_cnstrnt_decompsition}
\hat{\boldsymbol{\omega}}=\operatorname*{argmin}_{\boldsymbol{\omega}}\{||\mathbf{v}_i-\mathbf{D}\boldsymbol{\omega}||_2^2+\lambda g(\boldsymbol{\omega})\},
\end{equation}
where $\lambda$ is a positive hyperparameter and $g(\boldsymbol{\omega})$ the regularization function. 
By using $\hat{\boldsymbol{\omega}}$, we can estimate $\mathbf{v}_{0i}$ with equation (\ref{equ:dfv_rcvry}).

\subsubsection{Regularization}\label{sec:regularization}
$l_1$-norm and squared $l_2$-norm are two commonly selected regularization functions.
The $l_1$-norm leads to a sparse solution \cite{tropp2006just}\cite{donoho2008fast} with high computational cost,
while the squared $l_2$-norm has analytical solution and low computational complexity.
Both of them are studied in this paper.

For $l_1$-norm regularization, we name the approach SDBE\_L1 and equation (\ref{equ:general_cnstrnt_decompsition}) becomes
\begin{equation}\label{equ:l1norm_cnstrnt_decompsition}
\hat{\boldsymbol{\omega}}=\operatorname*{argmin}_{\boldsymbol{\omega}}\{||\mathbf{v}_i-\mathbf{D}\boldsymbol{\omega}||_2^2+\lambda ||\boldsymbol{\omega}||_1\}.
\end{equation}

Many fast implementations have been proposed for $l_1$-norm regularized LS estimate recently, such as the interior-point method \cite{kim2007interior} and DALM \cite{yang2013fast}. 
Even with these fast implementation, the $l_1$-norm regularized LS estimate is still computationally expensive.

For squared $l_2$-norm regularization, we name the approach SDBE\_L2 and equation (\ref{equ:general_cnstrnt_decompsition}) becomes
\begin{equation}\label{equ:l2norm_cnstrnt_decompsition}
\hat{\boldsymbol{\omega}}=\operatorname*{argmin}_{\boldsymbol{\omega}}\{||\mathbf{v}_i-\mathbf{D}\boldsymbol{\omega}||_2^2+\lambda ||\boldsymbol{\omega}||_2^2\},
\end{equation}
Equation (\ref{equ:l2norm_cnstrnt_decompsition}) has an analytical solution which can be easily derived as
\begin{equation}
\label{equ:l2norm_sltn}
\hat{\boldsymbol{\omega}}=\mathbf{P}\mathbf{v}_i ,
\end{equation}
where $\mathbf{P}=\left(\mathbf{D}^{T}\mathbf{D}+\lambda\mathbf{I}\right)^{-1}\mathbf{D}^{T}$.
Evidently, $\mathbf{P}$ is independent of $\mathbf{v}_i$ and can thus be calculated in advance as a procedure of the training process.
The computational complexity of (\ref{equ:l2norm_sltn}) is just proportional to the number of rows of $\mathbf{P}$, hence $\mathcal{O}(n)$.
Due to the low computational cost, the squared $l_2$-norm regularization is more suitable for the large-scale CD and OED.

\begin{figure}
	\begin{center}
		\includegraphics[width=1\linewidth]{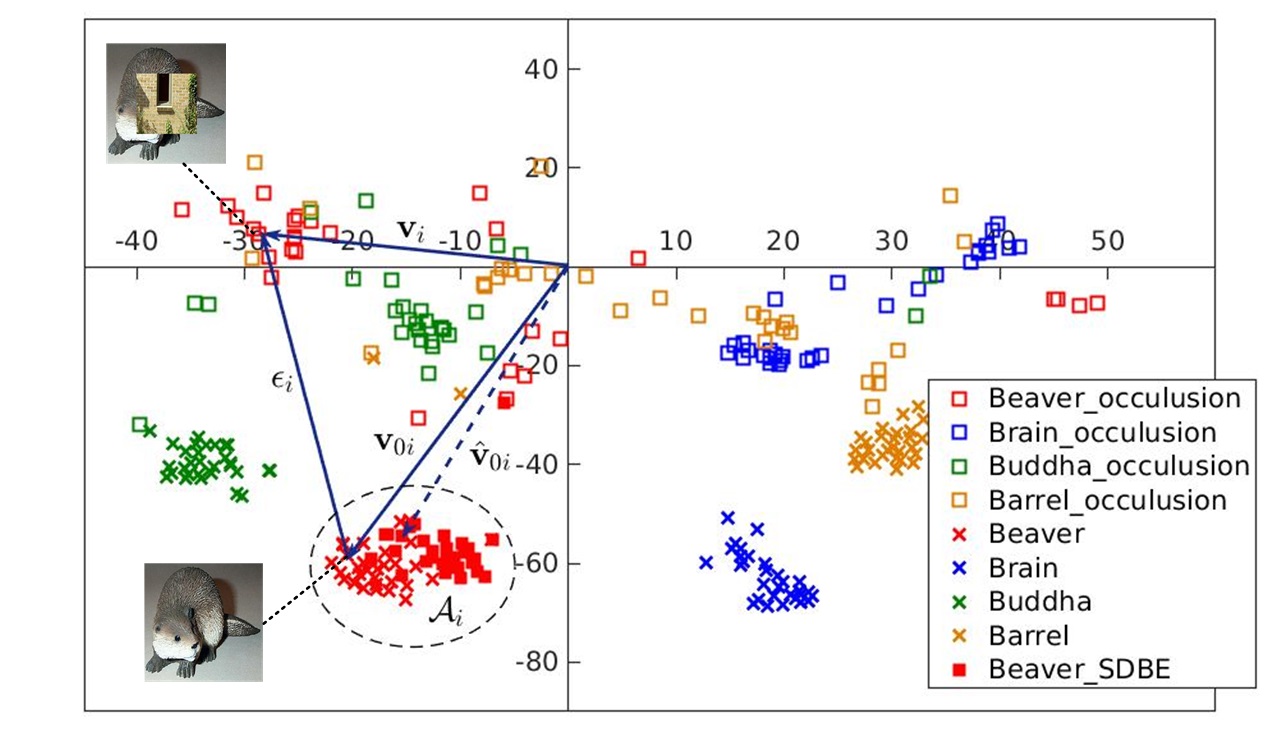}
	\end{center}	
	\caption{2-D illustration of DFVs of occluded images and occlusion-free images for Caltech-101 dataset. To make a clear presentation, only four categories each with 30 images are shown. The suffix "\_occlusion" denotes the DFV of occluded images and the suffix "\_SDBE" the DFV estimated by using the SDBE\_L2 approach. The base CNN is the ResNet-152 network and the SDBE\_L2 approach is employed to estimate the DFVs of the occlusion-free images. The t-SNE algorithm \cite{maaten2008visualizing} is used to map the DFVs from original $2048$-D to $2$-D for visualization. }
	\label{fig:SDBE_illustration}
\end{figure}

In \figurename{~\ref{fig:SDBE_illustration}}, to illustrate the effectiveness of the proposed SDBE to estimate the DFVs of the original occlusion-free images for the occluded images, we apply the SDBE\_L2 approach to the DFVs of the occluded "Beaver" images.
The experiment setting for the SDBE\_L2 approach is the same as the evaluation at $25\%$ occlusion ratio with the OED of case 1 in Section \ref{exp:estmt_accrcy}.
From the results, we can observe that most of the estimations (solid red square markers) fall into the vicinity of the class cluster of the "Beaver" category (red cross markers).

The implementation of SDBE-based classification is summarized in Algorithm~\ref{tab:lnnorm_algrthm}.
\begin{algorithm}
		\caption{The proposed SDBE\_L1 and SDBE\_L2 based classification algorithms}
		\label{tab:lnnorm_algrthm}
		{\bf Training Process}
			\begin{enumerate}
				\item\textbf{Input}: a set of training images and a set of extra image pairs.
				\item For each image, use CNNs to extract the DFV.
				\item Calculate $\mathbf{w}_{ij}$ by using (\ref{equ:wij_dfn}). 
				\item Construct $\mathbf{D}$ by using all $\mathbf{u}_{ij}$'s and $\mathbf{w}_{ij}$'s.
				\item Normalize each column of $\mathbf{D}$ to have unit $l_2$-norm (optional).
				\item For SDBE\_L2, calculate $\mathbf{P}=\left(\mathbf{D}^{T}\mathbf{D}+\lambda\mathbf{I}\right)^{-1}\mathbf{D}^{T}$.
				\item Train the classifier $\mathcal{C}$ with the column vectors of $\mathbf{A}$ (optional).
				\item \textbf{Output}: $\mathbf{D}$ (for SDBE\_L1) or $\mathbf{P}$ (for SDBE\_L2) and  $\mathcal{C}$.
			\end{enumerate}
		{\bf Testing Process}
			\begin{enumerate}
				\item \textbf{Input}: $\mathbf{D}$ (for SDBE\_L1) or $\mathbf{P}$ (for SDBE\_L2), $\mathcal{C}$ and a query image $\mathbf{y}_{i}$.	
				\item Use CNNs to extract the DFV of the query image and obtain $\mathbf{v}_i$.
				\item Normalize $\mathbf{v}_i$ to have unit $l_2$-norm (optional).				
				\item For SDBE\_L1, solve least-squares estimation problem (\ref{equ:l1norm_cnstrnt_decompsition}) and obtain $\hat{\boldsymbol{\omega}}=[\hat{\boldsymbol{\alpha}}^T\; \hat{\boldsymbol{\beta}}^T]^T$.\\
				For SDBE\_L2, calculate $\hat{\boldsymbol{\omega}}$ via 
				$\hat{\boldsymbol{\omega}} = \mathbf{P}\mathbf{v}_i$.
				\item Estimate $\mathbf{v}_{0i}$ by 
				$\hat{\mathbf{v}}_{0i}=\mathbf{A}\hat{\boldsymbol{\alpha}}$
				\item Normalize $\hat{\mathbf{v}}_{0i}$ to have unit $l_2$-norm (optional).
				\item Predict the class of $\hat{\mathbf{v}}_{0i}$ with the classifier $\mathcal{C}$.
				\item \textbf{Output}: The class of $\hat{\mathbf{v}}_{0i}$.
			\end{enumerate}			
\end{algorithm}	

For some classifiers, e.g., SVM, the input feature vector needs to have unit $l_2$-norm to achieve better performance, whereas, for other classifiers, e.g., the original softmax classifier of the ResNet-152 network, which is trained on unnormalized feature vectors, the input feature vector does not need to have unit $l_2$-norm. 
Therefore, the step (5) in the training phase and the step (3) and (6) in the testing phase are optional.
For a classifier requiring a normalized input, these steps will be conducted.
Otherwise, these steps will be skipped.
In addition, since in some application scenarios, for instance the experiments in Section \ref{exp:imagenet}, original softmax classifier of the CNN networks can be applied directly, the step (7) in the training phase is optional.

It is worthwhile to mention that the proposed scheme is generic for the classification of both occluded images and occlusion-free images, as demonstrated in Section \ref{exp:caltech_comprehensive} and \ref{exp:imagenet}.
The occlusion-free image corresponds to $\boldsymbol{\beta}=0$.

\section{Experiments}
\label{sec:Exps}
In this section, we extensively evaluate the proposed SDBE-based classification scheme on two publicly available datasets: Caltech-101 \cite{fei2007learning} and ImageNet \cite{ILSVRC15}.
The Caltech-101 dataset contains images of objects grouped into 101 categories, each with the number of images from 31 to 800, and embedded in cluttered backgrounds with different scales and poses.
The ImageNet is a comprehensive large-scale dataset.
A subset of ImageNet dataset, the ImageNet Large-Scale Visual Recognition Challenge 2012 (ILSVRC2012) \cite{ILSVRC15} classification dataset consisting of 1000 classes, is adopted for evaluation.
Each dataset is split into two set: \emph{class set} and \emph{extra set}.
The training images and the query images of occlusion-free version are drawn from the \emph{class set}. 
The occlusion-free images of the extra image pairs are drawn, unless otherwise specified, from the \emph{extra set}.

The ResNet-152 network \cite{he2016deep} pre-trained on the ILSVRC2012 dataset \cite{ILSVRC15} is adopted as the base CNN in the experiments (see supporting document for the experimental results for the pre-trained GoogLeNet\cite{szegedy2015going}).
The activations of the penultimate fully-connected layer, which is of 2048-D for the ResNet-152, are used as the DFV.

Several classifiers including NN, softmax, SVM, and the original softmax classifier of the ResNet-152 are adopted in different experiments to verify the improvement in classification accuracy. 
The LIBLINEAR implementation~\cite{fan2008liblinear} of $l_2$-regularized $l_2$-loss linear SVM is adopted for the SVM classifier in all the experiments and the penalty parameter of the linear SVM is selected from a grid set $\Theta=\{2^{-15},\dots,2^0,\dots,2^{15}\}$.

To match the input size of the base CNNs, all of the images are resized to $224\times 224$.
For the evaluations on the Caltech-101 dataset, the image is directly resized to $224\times 224$.
For the evaluations on the ImageNet dataset, the image is a center 224x224 crop from resized image with shorter side equal to 256 and the occluded image is generated by superimposing the occlusion patch onto the crop.
Since there are no suitable publicly available natural images datasets designed for the evaluation of the occluded image classification, we synthesize the occluded images by superimposing the occlusion patch on the resized images for evaluation.

For the experiments on the Caltech-101 dataset, the procedures of $l_2$ normalization (the step (5) in the training phase and the steps (3) and (6) in the testing phase in Algorithm \ref{tab:lnnorm_algrthm}) are adopted.
For the experiments on the ILSVRC2012 dataset, these procedures are skipped, since the original softmax classifier of the base CNN is not trained over the $l_2$ normalized DFVs.

In the experiments, 
the DALM \cite{yang2013fast} is adopted for the implementation of $l_1$-norm regularized LS minimization for the SDBE\_L1 approach.
The hyperparameter $\lambda$ is selected from a grid set $\boldsymbol{\Lambda}=\{10^{-6},\dots,0.5,1,\dots,10\}$. 
The MatConvNet~\cite{vedaldi15matconvnet} implementation of the ResNet-152 network is used for evaluation. 
Unless otherwise specified, the experiments are conducted on a PC with 16GB memory and an i7 CPU and without GPU acceleration.

\subsection{Evaluation on Caltech-101 dataset}\label{exp:caltech101dataset}
The Caltech-101 dataset excluding the "background" category is used to evaluated the basic properties of the proposed SDBE approaches.
In the experiments, the \emph{class set} includes $80$ categories with the names from "accordion" to "schooner" in alphabet order.
The remaining $21$ categories are treated as the \emph{extra set}.
The images of the training set consisting of $80$ categories are randomly drawn from each category of the \emph{class set}.
Except for the experiments in Section \ref{sec:size_cd}, each category of the training set contains $30$ images (the largest popular number of training images recommended on the website of the Caltech-101 dataset~\cite{caltech101website}).

For the evaluations at each occlusion ratio (except $0\%$ occlusion), the OED is formed by stacking the OEVs associated just with the testing occlusion ratio, e.g., for $25\%$ occlusion, only the OEVs associated with $25\%$ occlusion are included in the OED.
For the evaluations at $0\%$ occlusion, the same OED as that for $25\%$ occlusion is employed.

For clarity, the evaluated classification schemes are termed as a name combination of the proposed SDBE approach and the exploited classifer, e.g. "SDBE\_L1+SVM" indicates that the propsed SDBE\_L1 approach is followed by a linear SVM classifier.

\subsubsection{Estimation accuracy}\label{exp:estmt_accrcy}
This experiment is designed to show the estimation accuracy of the proposed SDBE approaches with respect to the occlusion ratio and the selection of occlusion-free extra images.
Two kinds of occlusion-free extra images are evaluated,
\begin{itemize}
	\item Case 1: the occlusion-free extra images are drawn from the \emph{extra set};
	\item Case 2: the occlusion-free extra images are drawn from the training images. 
\end{itemize}
For a fair comparison, the OEDs of both cases are kept in similar size.
We randomly draw $30$ images from each category of the \emph{extra set} in case $1$ and $8$ images from each category of the training images in case $2$.

\begin{figure}
	\centering
	\fontsize{8}{8}\selectfont
	\subfloat[]{
		\begin{minipage}{0.10\textwidth}
			\centering
			\includegraphics[width=1\linewidth]{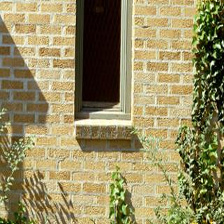}\\
			1
		\end{minipage}
		\begin{minipage}{0.10\textwidth}
			\centering
			\includegraphics[width=1\linewidth]{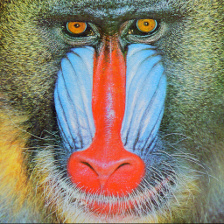}\\
			2
		\end{minipage}
		\begin{minipage}{0.10\textwidth}
			\centering
			\includegraphics[width=1\linewidth]{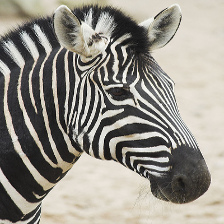}\\
			3
		\end{minipage}
		\begin{minipage}{0.10\textwidth}
			\centering
			\includegraphics[width=1\linewidth]{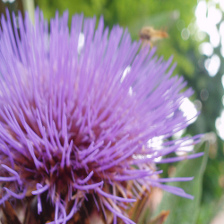}\\
			4
		\end{minipage}
	}\\
	\subfloat[]{
		\begin{minipage}{0.10\textwidth}
			\centering
			\includegraphics[width=1\linewidth]{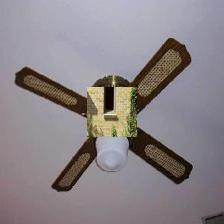}\\
			$5\%$
		\end{minipage}
		\begin{minipage}{0.10\textwidth}
			\centering
			\includegraphics[width=1\linewidth]{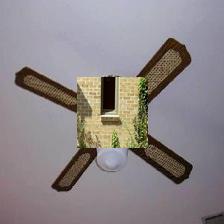}\\
			$10\%$
		\end{minipage}
		\begin{minipage}{0.10\textwidth}
			\centering
			\includegraphics[width=1\linewidth]{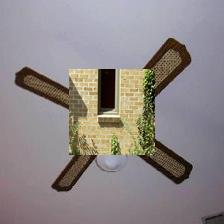}\\
			$15\%$
		\end{minipage}
		\begin{minipage}{0.10\textwidth}
			\centering
			\includegraphics[width=1\linewidth]{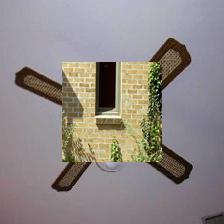}\\
			$20\%$
		\end{minipage}
		\begin{minipage}{0.10\textwidth}
			\centering
			\includegraphics[width=1\linewidth]{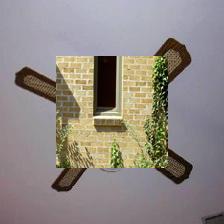}\\
			$25\%$
		\end{minipage}
		\begin{minipage}{0.10\textwidth}
			\centering
			\includegraphics[width=1\linewidth]{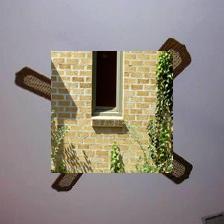}\\
			$30\%$
		\end{minipage}
		\begin{minipage}{0.10\textwidth}
			\centering
			\includegraphics[width=1\linewidth]{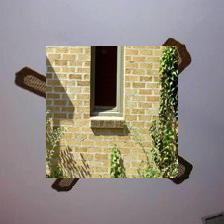}\\
			$35\%$
		\end{minipage}
		\begin{minipage}{0.10\textwidth}
			\centering
			\includegraphics[width=1\linewidth]{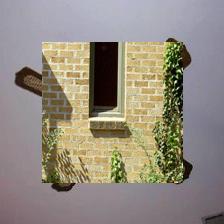}\\
			$40\%$
		\end{minipage}
		\begin{minipage}{0.10\textwidth}
			\centering
			\includegraphics[width=1\linewidth]{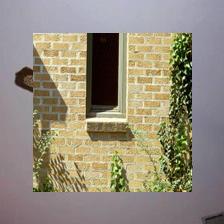}\\
			$50\%$
		\end{minipage}
	}
	\\
	\subfloat[]{
		\begin{minipage}{0.48\textwidth}
			\includegraphics[width=1\linewidth]{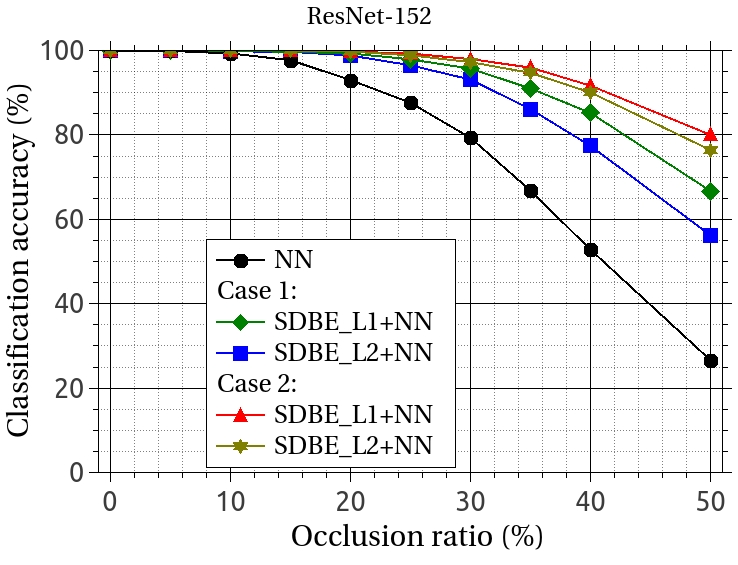}
		\end{minipage}
	}
	\subfloat[]{
		\begin{minipage}{0.48\textwidth}
			\includegraphics[width=1\linewidth]{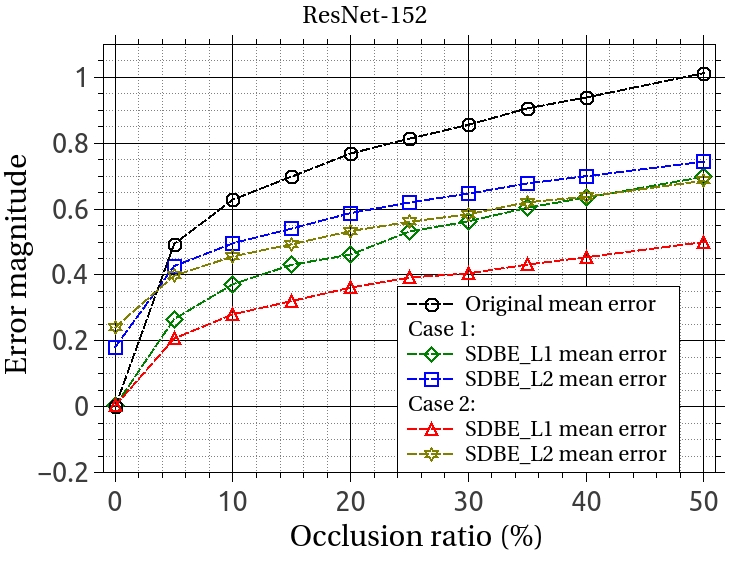}
		\end{minipage}
	}	
	\caption{Comparison of classification accuracies and estimation errors with respect to occlusion ratio for different classification schemes. (a) Four occlusion patches used in the experiment. (b) The examples of occluded images with various occlusion ratios. 
		(c) Classification accuracies and (d) estimation errors for the pre-trained ResNet-152 base CNN.
		SDBE\_L1 mean error and SDBE\_L2 mean error are the mean estimation errors of the proposed SDBE\_L1 and SDBE\_L2, respectively, and the original mean error is the mean distance between $\mathbf{v}_i$ and $\mathbf{v}_{0i}$ for the query images occluded with testing occlusion ratio.
	}
	\label{fig:Inc_cntr_occ}
\end{figure}

In the experiment, all of the occluded images are contaminated at the respective image centers.
Four occlusion patches collected from the outside of the \emph{class set} and extra set, as shown in \figurename{~\ref{fig:Inc_cntr_occ}}(a), are used to synthesize the occluded images.
Only the first one is used to produce the query images, while all of four occlusion patches are employed to generate the occluded extra images.
Therefore, in the OED, the last three occlusion patches in \figurename{~\ref{fig:Inc_cntr_occ}}(a) are 
regarded as the interferences.
Eventually, for each evaluation, the OED includes $30*21*4=2520$ OEVs for case 1 and $8*80*4=2560$ OEVs for case 2.

In order to eliminate the impact of the error caused by \emph{class subspace} approximation and assess the estimation error directly, the occluded query images are synthesized by corrupting the training images.
In the experiment, the estimation error is measured with the Euclidean distance, i.e., $||\mathbf{v}_{0i}-\hat{\mathbf{v}}_{0i}||_2$.
An NN classifier is adopted to determine the category of the query image.
The best results with respect to $\lambda\in\boldsymbol{\Lambda}$ for each occlusion ratio are shown in \figurename{~\ref{fig:Inc_cntr_occ}}. 

From \figurename{~\ref{fig:Inc_cntr_occ}}, we have the following observations.
\begin{enumerate}
	\item For both of the cases, the mean estimation errors of the SDBE\_L1 and SDBE\_L2, except for zero occlusion, are much smaller than the original mean errors.
	\item For both of the cases, except for zero occlusion, the SDBE\_L1 and SDBE\_L2, though with pretty large mean estimation errors, significantly improve the classification accuracy, especially at the high occlusion ratios.
	\item The SDBE\_L2 in case 2 achieves higher classification accuracy with larger estimation error than the SDBE\_L1 in case 1.
	\item The SDBE\_L1 leads to much smaller estimation error and much higher classification accuracy than the SDBE\_L2 for each case. 
	\item For zero occlusion, the classification scheme with either the SDBE\_L1 or the SDBE\_L2 achieves the same classification accuracy as that without the proposed SDBE approaches. The SDBE\_L1 can keep the DFV almost intact for zero occlusion.
	\item The OED of case 2 achieves higher classification accuracy than the OED of case 1.
\end{enumerate}

The first observation demonstrates the effectiveness of the proposed SDBE approaches to estimate the DFV of the original occlusion-free image.
The second and third show that the target of classification differs from that of signal recovery.
In classification, the estimation with larger error can be acceptable as long as it gives rise to a better classification result.

For the fourth, it is mainly because the CD $\mathbf{A}$ contains $\mathbf{v}_{0i}$ such that the sparsest solution, $\boldsymbol{\alpha}$ with just one nonzero entry corresponding to $\mathbf{v}_{0i}$, becomes the best solution.
Nevertheless, in a more frequently encountered situation that $\mathbf{A}$ does not include $\mathbf{v}_{0i}$, the difference between the SDBE\_L1 and the SDBE\_L2 can be very small, even unnoticeable, in terms of classification accuracy, given a powerful classifier.

The fifth observation indicates that the proposed SDBE\_L1 and SDBE\_L2 are generic for 
the classification of both occluded query images and occlusion-free query images (corresponding to zero occlusion). 
This property is also demonstrated in the following experiments.

The sixth observation is because the OEVs generated by using the images drawn from the categories of the query images are related more tightly to the OEVs of the occluded query images.
Nevertheless, case 1 is more frequently encountered in practice, since the occluded images of specified categories are more difficult to collect than those of arbitrary categories.
Furthermore, since the performance of case 2 is obviously superior to that of case 1, it is reasonable to assume that a greater improvement is able to be achieved by substituting the OED of case 2 for the OED of case 1.
Therefore, in the following experiment, we just consider the OED of case 1. 

\subsubsection{Hyperparameter $\lambda$}\label{exp:hyprprmtr}
This experiment is designed to show the performance variations of the proposed SDBE approaches with respect to the hyperparameter $\lambda$ in terms of classification accuracy.  
In the experiment, all of the settings except for the query images and the classifier are the same as case 1 in Section \ref{exp:estmt_accrcy}.
A maximum of 50 images for each category are randomly drawn to synthesize the occluded query images from the \emph{class set} excluding the training images.
Since the occluded query images are synthesized by using the images outside the training set, two more powerful classifiers, linear SVM and softmax, are employed for evaluation.
Two occlusion ratios, $15\%$ and $25\%$, are tested in the experiment.

\begin{figure}
	\centering
	\fontsize{8}{8}\selectfont
	\includegraphics[width=0.95\linewidth]{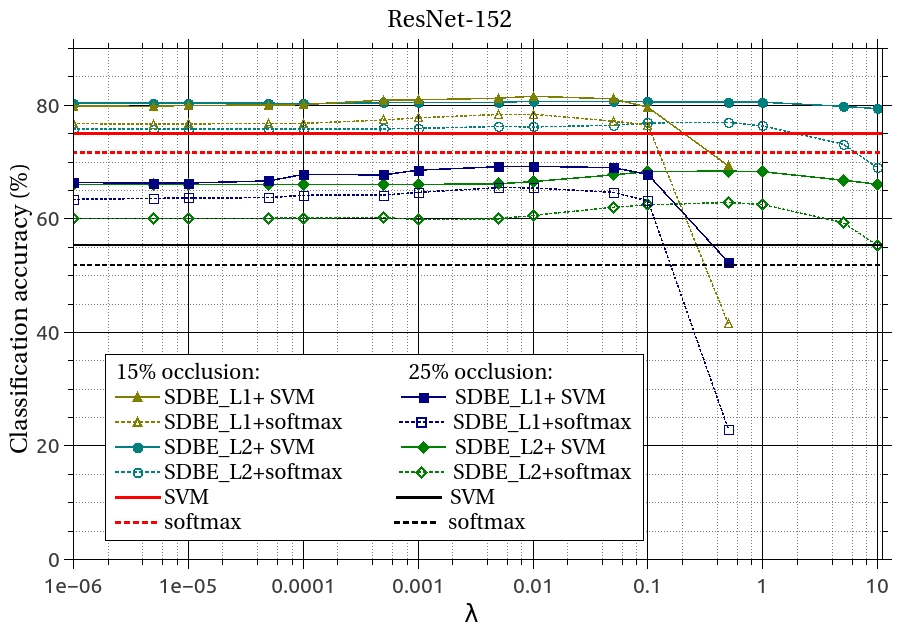}
	\caption{Classification accuracy with respect to $\lambda$ for ResNet-152. The classification results of linear SVM and softmax without the proposed SDBE methods are also shown for comparison.}
	\label{fig:hyperpara_lambda}
\end{figure}
The experimental results are shown in \figurename{~\ref{fig:hyperpara_lambda}}.
From the results, we can observe that the performance variations are pretty small in the ranges of $[10^{-6}, 0.1]$ for the SDBE\_L1 and $[10^{-6}, 1]$ for the SDBE\_L2.
In addition, as can be seen, for distinct occlusion ratios, the $\lambda$'s for best results are very similar.
Therefore, the proposed SDBE approaches are not very sensitive to $\lambda$ with respect to occlusion ratio, which can extend the application scenarios of the proposed SDBE approaches. 

\subsubsection{OED Size}\label{sec:size_oed}
This experiment is designed to show the classification accuracy with respect to the size setting of the OED.
In the experiment, the settings for the training images, occlusion patches, and query images are the same as those used in Section \ref{exp:hyprprmtr}.
The occlusion-free extra images are randomly drawn from those used in Section \ref{exp:hyprprmtr} according to the size setting of the OED. 
Suppose that the occlusion pattern used to contaminate the query images is the $j$th occlusion pattern.

The best experimental result with respect to $\lambda$ for each setting is shown in \tablename{~\ref{tab:cltch101_oed_setting_ResNet}}, where $p_j$, $N_c$, and $N_I$ denote the number of the OEVs associated with the $j$th occlusion pattern, the number of image categories, and the number of occlusion-free extra images per category, respectively, and $p_j=N_c*N_I$.
Since three occlusion patches involved in the construction of OED are taken as the interferences, the overall number of the OEVs in the OED is four times $p_j$.

\begin{table}
	\renewcommand{\arraystretch}{1.3}	
	\caption{Comparison of OED settings in terms of classification accuracy (\%) for ResNet-152. OR: occlusion ratio.}
	\centering
	\fontsize{8}{8}\selectfont
	\begin{tabular}{c|c|c|c|c|c|c|c}
		\hline
		OR&\multicolumn{3}{c|}{OED} & \multicolumn{2}{c|}{SDBE\_L1+} &\multicolumn{2}{c}{SDBE\_L2+} \\
		\cline{2-8}
		 & $p_j$&$N_c$ & $N_I$& SVM &softmax &SVM & softmax \\
		\hline
		& $21$&$1$ & $21$& 80.3 &76.9 & 79.6& 76 \\
		\cline{2-8}
		 & $21$&$21$ & $1$& 80 &76.7 &80.1 & 76.9 \\
		 \cline{2-8}
			& $105$&$5$ & $21$& 80.7 &77.9 &80.6 & 77 \\
			\cline{2-8}
	  $15\%$&$105$&$7$ & $15$& 81 &78 &80.9 & 77 \\
			\cline{2-8}
			& $105$&$15$ & $7$& 80.9 &\textbf{79.3} &81.0 & 77.4 \\
			\cline{2-8}
			& $105$&$21$ & $5$& 80.9 &78.2 &80.7 & \textbf{77.8} \\
		 \cline{2-8}
		 & $210$&$21$ & $10$& 81.2 &78.3 &\textbf{81.1} & 77.5 \\
		 \cline{2-8}
		 & $630$&$21$ & $30$& \textbf{81.6} &78.4 &80.7 & 77 \\
		 \hline
		 & $21$&$1$ & $21$& 66.3 &62.5 &66.5 & 60.3 \\
		 \cline{2-8}
		 & $21$&$21$ & $1$& 67.1 &64.1 &66.1 & 62 \\
		 \cline{2-8}
		 & $105$&$5$ & $21$& 67.4 &64 &66.9 & 62.1 \\
		 \cline{2-8}
	  $25\%$&$420$&$7$ & $15$& 67.8 &64.8 &67.4 & 62.6 \\
		 \cline{2-8}
		 & $105$&$15$ & $7$& 67.9 &63.6 &68.1 & 62.9 \\
		 \cline{2-8}
		 & $105$&$21$ & $5$& 67.7 &63.3 &67.3 & \textbf{63.6} \\
		 \cline{2-8}
		 & $210$&$21$ & $10$& 68.7 &\textbf{66} &68 & \textbf{63.6} \\
		 \cline{2-8}
		 & $630$&$21$ & $30$& \textbf{69.3} &\textbf{66} &\textbf{68.4} & 62.8 \\
		 \hline
	\end{tabular}
	\label{tab:cltch101_oed_setting_ResNet}
\end{table}

From \tablename{~\ref{tab:cltch101_oed_setting_ResNet}}, we can observe that the classification accuracy increases primarily with $p_j$, e.g., $p_j=21,105,210$ and gradually approaches to stable, e.g., $p_j=210,630$,
while, the diversity of the categories does not considerably affect the classification accuracy, e.g., the differences between the classification accuracies are smaller than $1.3\%$ and $1.1\%$ for $N_c=1, 21$ and $N_c=5,7,15,21$, respectively.
The saturation with respect to $p_j$ indicates that the linear span of the OEVs is not effectively extended by excessive OEVs.
The small discrepancy with respect to the diversity of the categories shows that the relative positions between the DFVs of the occluded images and the DFVs of the original occlusion-free images are similar for different categories in the \emph{extra set}.
Therefore, in practice, we do not need to collect too much extra image pairs for each occlusion pattern and less consideration is required on the diversity of the categories when collecting the extra image pairs.

\subsubsection{CD Size}\label{sec:size_cd}
\begin{figure}
	\centering
	\fontsize{8}{8}\selectfont
	\subfloat[]{
		\begin{minipage}{0.5\textwidth}
			\includegraphics[width=1\linewidth]{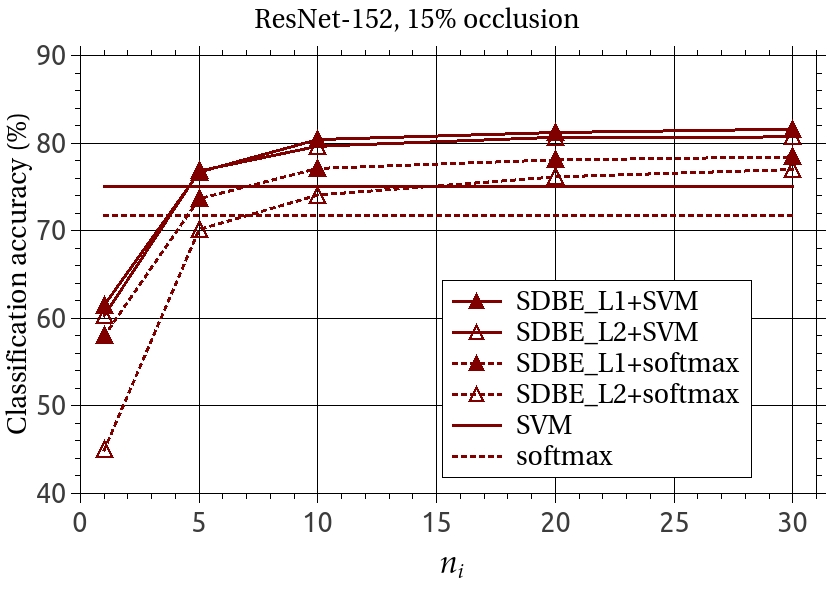}
		\end{minipage}
	}
	\subfloat[]{
		\begin{minipage}{0.5\textwidth}
			\includegraphics[width=1\linewidth]{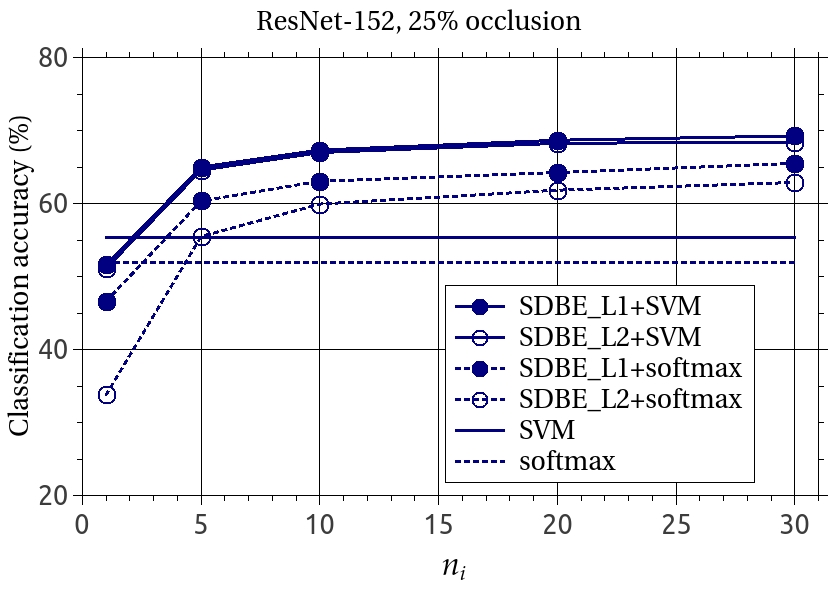}
		\end{minipage}
	}
	\caption{Comparison of classification accuracies with respect to the CD size for different classification schemes with the ResNet-152 network. Each training category contains the same number of images $n_i$. Two occlusion ratios, (a) $15\%$ and (b) $25\%$, with the center contamination are evaluated.}
	\label{fig:exp_size_cd}
\end{figure}

This experiment is designed to show the influence of the size of the CD on the classification accuracy.
In the experiment, two occlusion ratios, $15\%$ and $25\%$, are considered for evaluation.
For the evaluations at each occlusion ratio, the OED, query images, and training images used to train the classifiers are the same as those used for the evaluations at each respective occlusion ratio in Section \ref{exp:hyprprmtr}.
The training images used to construct the CD are randomly drawn from the training images used to train the classifiers according to the size of the CD.
The best classification results with respect to $\lambda$ for each occlusion ratio are shown in \figurename{~\ref{fig:exp_size_cd}}.

From \figurename{~\ref{fig:exp_size_cd}}, we can learn that the CD of extremely small size, e.g., $n_i=1$, leads to a degradation in classification accuracy due to the large error in the approximation of the \emph{class subspace} induced by the extremely under-sample of the CD over the class space.
Fortunately, the classification accuracies, however, increase rapidly with the size of CD, e.g., for $n_i=5$, the classification accuracies of the schemes with the proposed SDBE approaches are much higher than those without, and eventually, the classification accuracies saturate, e.g., for $n_i\geqslant 10$ , the classification accuracies increase very small with the size of the CD.

As analyzed in Section \ref{sec:regularization}, the computational load of the proposed SDBE approaches will rise with the sizes of the CD and the OED. 
So, the experiments in Section \ref{sec:size_oed} and \ref{sec:size_cd} reveal that, in practical applications, we can make an optimal trade-off between the classification accuracy and the sizes of the CD and the OED to fulfill a classification accuracy requirement within a limited computational cost.

\subsubsection{Comprehensive classification}\label{exp:caltech_comprehensive}
This experiment is designed to evaluate the proposed SDBE approaches in a comprehensive situation.

\begin{figure*}
	\centering
	\fontsize{6}{8}\selectfont	
	\includegraphics[width=0.8\linewidth]{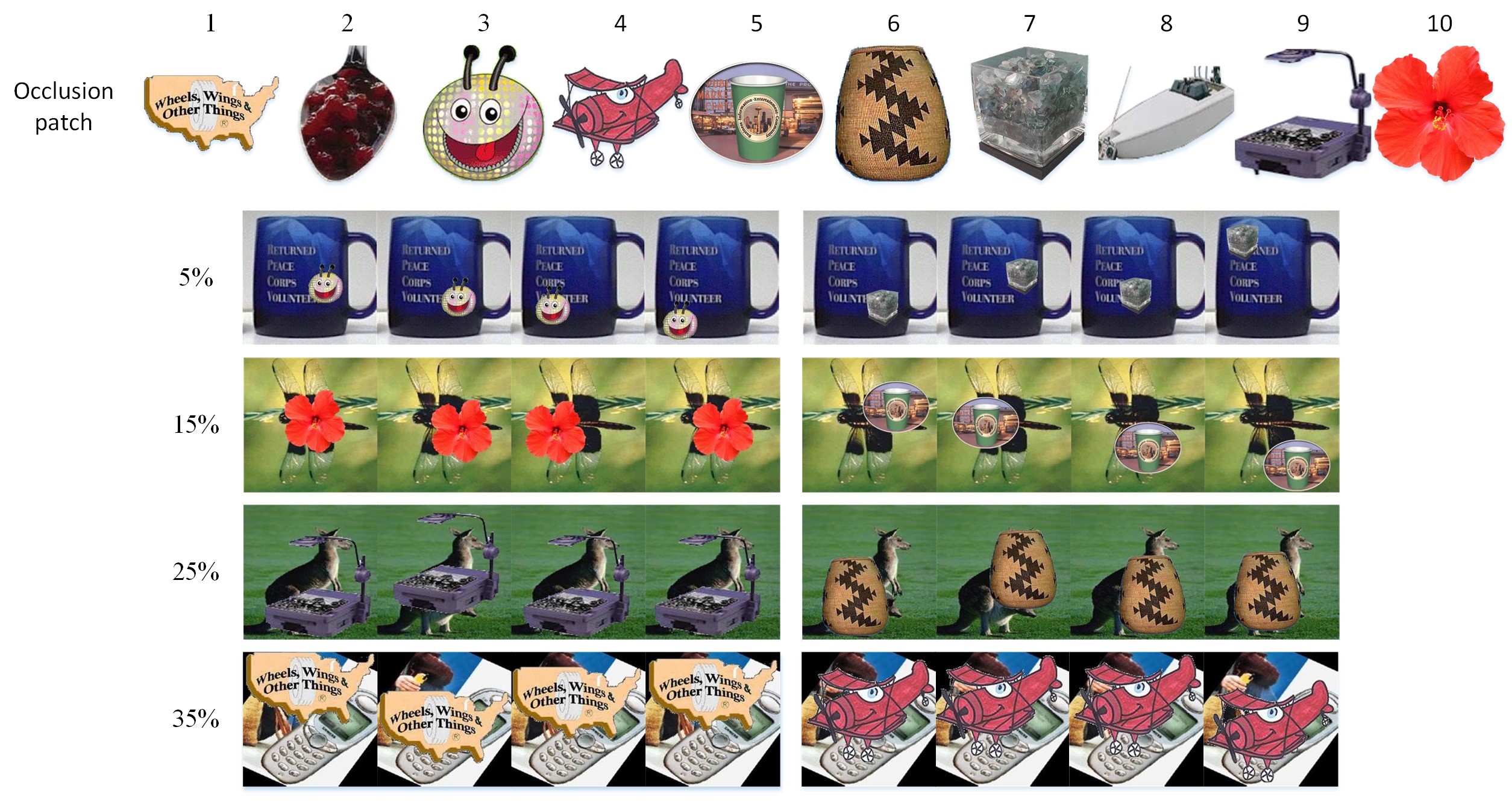}
	\caption{Examples of occlusion patches and occluded query images used in the comprehensive experiment on Caltech-101 dataset. For each occlusion ratio, the examples of two occlusion patches are presented so as to show that the random occlusion positions for each occlusion patch are independently sampled.	
	}
	\label{fig:cltch_cmprhnsv_example_images}
\end{figure*}

In the experiment, for the evaluation at each occlusion ratio, $40$ occlusion patterns, which involve ten occlusion patches manually segmented from the images in the background category, as shown in \figurename{~\ref{fig:cltch_cmprhnsv_example_images}}, each with $4$ random occlusion positions, are employed to construct the OED and synthesize the occluded query images.

For the OED, we set $N_c=21$ and $N_I=5$.
The occlusion-free extra images are randomly drawn from each category of the \emph{extra set}.
Then, we have $21*5*40=4200$ OEVs in the OED for the evaluation at each occlusion ratio.
The occlusion-free query images used to synthesize the occluded query images are the same as those used in Section \ref{exp:hyprprmtr}. 
The examples of the occluded query images are shown in \figurename{~\ref{fig:cltch_cmprhnsv_example_images}}. 

The regularization parameter of the linear SVM and the hyperparameter $\lambda$ of the proposed SDBE approaches are determined by using a $3$-folder cross-validation method.

For a thorough comparison, three conventional classification schemes, of which the classifiers are trained with different training sets, are evaluated.
The first, denoted as 'orginal', is a case frequently encountered in practice, i.e., the training set merely consists of the occlusion-free images, which is the same as those for the proposed SDBE approaches.
The second and third contain the occlusion-free training images used for the first scheme plus occluded training images.
The second, denoted as 'org-full', corresponds to an ideal situation that one can be aware of the information about all occlusion patterns for each occlusion-free training image, i.e., the occluded training images are generated by contaminating each occlusion-free training image with each occlusion pattern.
Such an ideal situation is, however, rarely encountered in practical applications due to the difficulty in collecting real occluded images.
The third, denoted as 'org-partial', employs a training set containing the same number of occluded images as that in the \emph{extra set} used for the proposed SDBE approaches, i.e., the occluded training images are produced by applying each occlusion pattern to a small subset of occlusion-free training images, which consists of $21$ categories each with $5$ images randomly drawn from the occlusion-free training set.
Therefore, in the third case, the classifiers can acquire the same amount of knowledge about the occlusion patterns as the proposed SDBE approaches.
Eventually, in the second and the third case, the training sets are composed of $96000$ and $4200$ occluded training images, respectively.

From \tablename{~\ref{tab:Caltech-101_comprehensive}}, we observe that the 'org-full' schemes achieve the best classification results for the occluded images, but with over $1.3\%$ loss in classification accuracy for the occlusion-free images.
While, the 'org-partial' schemes only show small improvements at the large occlusion ratios ($25\%$ and $35\%$ occlusions for linear SVM classifier and $35\%$ occlusion for softmax classifier).
These results are attributed to inconsistency between the statistics of the training samples and query samples.
For instance, in the 'org-full' schemes, the training sets include much more occluded images than the occlusion-free images and a compromise is made between the occlusion-free and the occluded training images for the classification boundary.
Consequently, for the evaluations at the zero occlusion, where no occluded images are involved in the query images, the performance is degraded.

\begin{table*}
	\renewcommand{\arraystretch}{1.4}
	\centering
	\fontsize{8}{8}\selectfont
	\begin{tabular}{c|l|c|c|c|c|c|c|c|c|c|c}
		\hline
		\multicolumn{2}{c|}{OR}                                                               & \multicolumn{2}{c|}{0\%}                            & \multicolumn{2}{c|}{5\%}                            & \multicolumn{2}{c|}{15\%}                           & \multicolumn{2}{c|}{25\%}                           & \multicolumn{2}{c}{35\%}                           \\ \hline
		&         & \multicolumn{1}{c|}{$R$(\%)} & \multicolumn{1}{c|}{$E$} & \multicolumn{1}{c|}{$R$(\%)} & \multicolumn{1}{c|}{$E$} & \multicolumn{1}{c|}{$R$(\%)} & \multicolumn{1}{c|}{$E$} & \multicolumn{1}{c|}{$R$(\%)} & \multicolumn{1}{c|}{$E$} & \multicolumn{1}{c|}{$R$(\%)} & \multicolumn{1}{c}{$E$} \\ \hline
		\multirow{2}{*}{orginal}                                                                   & SVM     & 94.89                      & 0                      & 91.63                      & 0.444                  & 74.25                      & 0.689                  & 50.17                      & 0.827                  & 30.45                      & 0.921                  \\ \cline{2-12}
		& softmax & 93.50                      & 0                      & 89.65                      & 0.444                  & 72.30                      & 0.689                  & 48.84                      & 0.827                  & 29.37                      & 0.921                  \\ \hline
		\multirow{2}{*}{org-partial} & SVM     & 94.36                      & 0                      & 90.76                      & 0.444                  & 71.95                      & 0.689                  & 51.15                      & 0.827                  & 36.53                      & 0.921                  \\ \cline{2-12} 
		& sofmax  & 93.58                      & 0                      & 89.88                      & 0.444                  & 68.75                      & 0.689                  & 44.08                      & 0.827                  & 30.58                      & 0.921                  \\ \hline
		\multirow{2}{*}{org-full}  & SVM     & 93.58                      & 0                      & 93.11                      & 0.444                  & 88.09                      & 0.689                  & 81.65                      & 0.827                  & 74.11                      & 0.921                  \\ \cline{2-12} 
		& softmax & 92.15                      & 0                      & 91.93                      & 0.444                  & 86.35                      & 0.689                  & 79.43                      & 0.827                  & 71.05                      & 0.921                  \\ \hline
		\multirow{2}{*}{SDBE\_L1+}                                                    & SVM     & 95.02                      & 0.35                   & 92.68                      & 0.483                  & 85.51                      & 0.569                  & 74.71                      & 0.636                  & 62.66                      & 0.683                  \\ \cline{2-12} 
		& softmax & 93.12                      & 0.495                  & 90.86                      & 0.483                  & 83.12                      & 0.569                  & 71.58                      & 0.636                  & 58.09                      & 0.686                  \\ \hline
		\multirow{2}{*}{SDBE\_L2+}                                                    & SVM     & 95.23                      & 0.351                  & 92.56                      & 0.481                  & 84.72                      & 0.573                  & 73.56                      & 0.645                  & 61.88                      & 0.696                  \\ \cline{2-12} 
		& softmax & 93.37                      & 0.358                  & 90.01                      & 0.484                  & 81.42                      & 0.58                   & 69.33                      & 0.648                  & 55.93                      & 0.696                  \\ \hline
	\end{tabular}
	\caption{Comparison of classification accuracies and mean errors with respect to occlusion ratio for different schemes on Caltech-101 dataset. OR: occlusion ratio. $R$: classification accuracy. $E$: magnitude of mean error. For the proposed SDBE\_L1 and SDBE\_L2, mean error is the mean estimation error, and for the conventional schemes, original, org-partial, and org-full, mean error is the mean distance between $\mathbf{v}_i$ and $\mathbf{v}_{0i}$. The pre-trained ResNet-152 network is adopted as the base CNN.}
	\label{tab:Caltech-101_comprehensive}
	\vspace{-5pt}
\end{table*}

On the contrary, by estimating the DFVs of the occlusion-free images at the testing phase, the proposed SDBE approaches do not change the statistic of the training set and thus, maintain the performance for occlusion-free images and drastically boost the classification accuracy for the occluded images.   
For instance, in \tablename{~\ref{tab:Caltech-101_comprehensive}}, at $35\%$ occlusion, the SDBE\_L1 achieves $32.21\%$ and $28.72\%$ increase over the 'original' and $26.13\%$ and $27.51\%$ increase over the 'org-partial' in classification accuracy for the linear SVM and softmax classifier, respectively.
In the meantime, the SDBE\_L1 merely introduce slight changes ($0.13\%$ increase for linear SVM and $0.38\%$ reduction for softmax) under the zero occlusion.
Therefore, for the situation that only a small-scale set of occluded images are available in the training phase, the proposed SDBE-based classification scheme is a better choice than the conventional classification scheme.

By comparing the results for the SDBE\_L1 and SDBE\_L2 in \tablename{~\ref{tab:Caltech-101_comprehensive}}, we can learn that the SDBE\_L1 usually achieves smaller estimation errors and better classification accuracies.
However, with a powerful classifier, such as the linear SVM, which is better than the softmax for small-scale training sets, the SDBE\_L2 achieves similar performance as the SDBE\_L1.
For instance, at the $35\%$ occlusion, the estimation error gaps between the SDBE\_L1 and SDBE\_L2 for the linear SVM and softmax are almost the same, while, the classification accuracy gap for the linear SVM is reduced from $2.19\%$ for the softmax to $0.78\%$.
The major drawback of the SDBE\_L1 as mentioned in the above sections is the high computational complexity. 

\subsection{Evaluation on ImageNet dataset}\label{exp:imagenet}
The experiments in this section are designed to evaluate the proposed SDBE approaches on the ILSVRC2012 classification dataset.
The original "fc1000" and "prob" layers of the pre-trained ResNet-152 network, which actually constitute a softmax classifier, are adopted as the classifier in the experiments.

In the experiments, the \emph{class set} consists of $900$ categories randomly drawn from all $1000$ categories and the \emph{extra set} contains $20$ categories randomly drawn from the remaining $100$ categories.
The occlusion patches are segmented from the images of the rest $80$ categories.
The training images used to produce the CD are randomly drawn from the ILSVRC2012 training images of each category in the \emph{class set}.
Two occlusion ratios, $10\%$ and $20\%$, are tested in the experiments.
Unlike the experiments on the Caltech-101 dataset, only one OED is constructed for the testings at different occlusion ratios in this case.
The occlusion-free extra images used to construct the OED have $20$ categories each with $5$ images randomly drawn from the images of the \emph{extra set}.

Two experiments are conducted in this section.
The first experiment focuses on demonstrating the improvements of the SDBE\_L1 and SDBE\_L2 over the original network.
Due to the rapid increase in computational complexity with the sizes of the CD and OED for the SDBE\_L1, the CD and OED of small size are adopted to make the evaluation time tolerable in the first experiment.
The CD consists of $900$ categories each with $5$ DFVs and the OED involving $16$ occlusion patterns is made up of $1600$ OEVs associated with $20$ categories each with $80$ OEVs.
The $16$ occlusion patterns are produced with $4$ occlusion patches, as shown in \tablename{~\ref{tab:ILSVRC_exp1}}, each with two random occlusion positions at each of the two occlusion ratios.

\begin{figure*}
	\centering
	\fontsize{8}{8}\selectfont
	\includegraphics[width=0.8\linewidth]{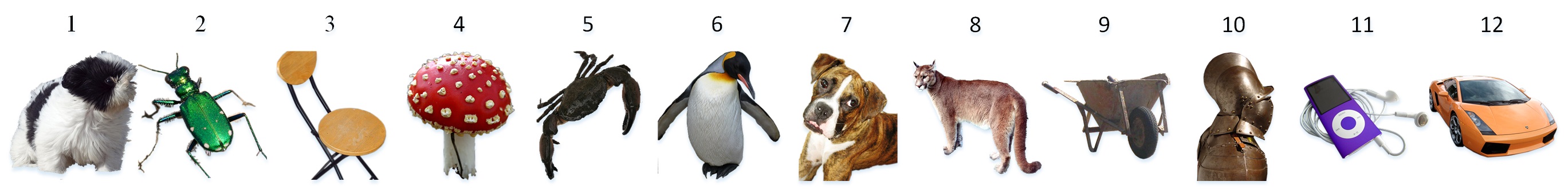}\\
	(a)\\
	\includegraphics[width=0.8\linewidth]{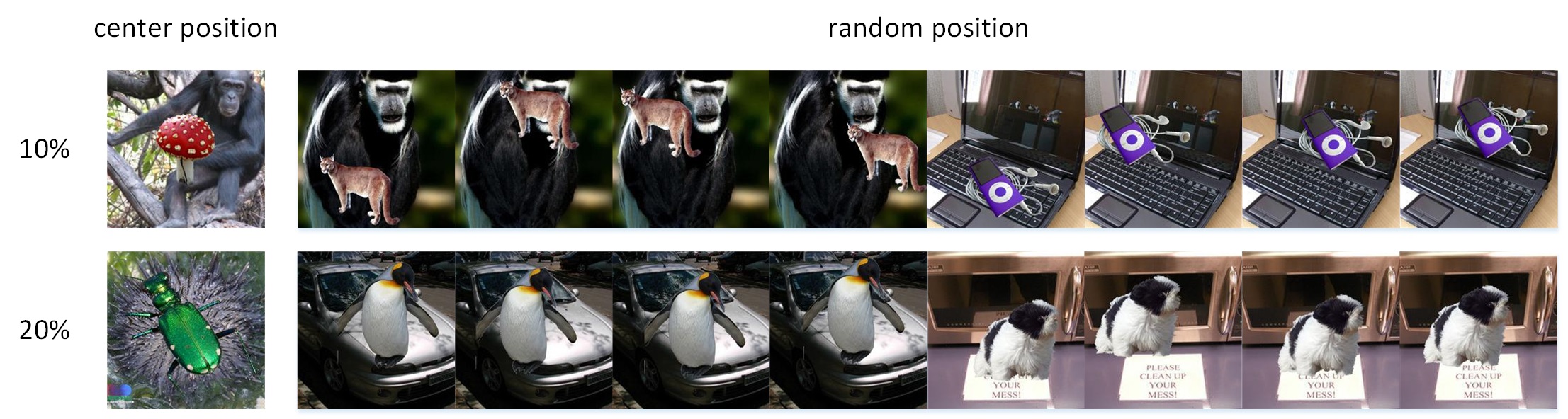}\\
	(b)
	\caption{Examples of the occlusion patches and occluded query images used in the second experiment on ILSVRC dataset. (a) Occlusion patches. (b) examples of the occluded query images. For each occlusion ratio, the examples of two occlusion patches are presented for the random occlusion positions so as to show that the random occlusion positions of each occlusion patch are independently sampled.  
	}
	\label{fig:massive_patchs_list}
\end{figure*}

The second experiment aims at showing the scalability of the SDBE\_L2 to handle a large number of occlusion patterns, and thus, only the original network and SDBE\_L2 are evaluated in the second experiment.
The number of the occlusion patterns is extended to $120$ and the number of DFVs in the CD for each category to $20$.
The $120$ occlusion patterns are constituted by $12$ occlusion patches including $4$ occlusion patches used in the first experiment, as shown in \figurename{~\ref{fig:massive_patchs_list}}, each with a center and four random occlusion positions at each of the two occlusion ratios.
Eventually, the CD and OED are composed of $18000$ DFVs and $12000$ OEVs, respectively.

\begin{table}[]
	\renewcommand{\arraystretch}{1.4}
	\centering
	\fontsize{8}{8}\selectfont
	\includegraphics[width=0.7\linewidth]{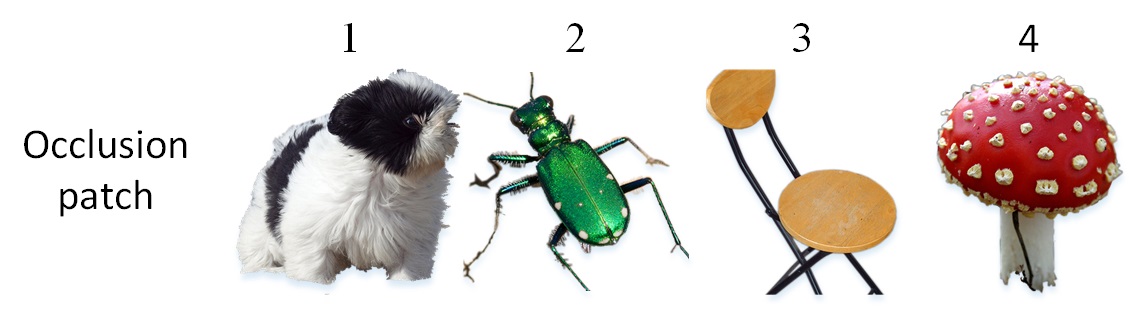}\\
	\begin{tabular}{c|c|c|c|c|c|c}
		\hline
		\multirow{2}{*}{OR}                     & \multirow{2}{*}{method} & \multicolumn{4}{c|}{occlusion patch} & \multirow{2}{*}{Avg.} \\ \cline{3-6}
		&                         & 1       & 2       & 3       & 4      &                       \\ \hline
		\multirow{3}{*}{0\%} & Original                &\textemdash&\textemdash&\textemdash&\textemdash&    75.39             \\ \cline{2-7} 
		& SDBE\_L1                &\textemdash&\textemdash&\textemdash&\textemdash&    74.92              \\ \cline{2-7} 
		& SDBE\_L2                &\textemdash&\textemdash&\textemdash&\textemdash&    74.89              \\ \hline
		\multirow{3}{*}{10\%}                   & Original                & 53.56   & 11.08   & 56.32   & 55.39  & 44.09                 \\ \cline{2-7} 
		& SDBE\_L1                & 64.60   & 52.32   & 62.70   & 64.53  & 61.04                 \\ \cline{2-7} 
		& SDBE\_L2                & 64.60   & 52.32   & 62.69   & 64.50  & 61.03                 \\ \hline
		\multirow{3}{*}{20\%}                   & Original                & 29.70   & 1.77    & 36.89   & 39.42  & 26.94                 \\ \cline{2-7} 
		& SDBE\_L1                & 57.84   & 33.90   & 50.59   & 54.50  & 49.21                 \\ \cline{2-7} 
		& SDBE\_L2                & 57.84   & 33.90   & 50.59   & 54.48  & 49.20                 \\ \hline
	\end{tabular}
	\caption{Comparison of classification accuracy (\%) for different classification schemes on ILSVRC2012 dataset. Four occlusion patches used in the experiment are shown above the table (OR: occlusion ratio). The pre-trained ResNet-152 network is adopted as the base CNN.}
	\label{tab:ILSVRC_exp1}
\end{table}

{ The validation images of the ILSVRC2012 dataset in the \emph{class set} are adopted as the zero occlusion query images for both experiments. In the first experiment, to reduce the workload for the evaluation of the occluded query images, a subset of the zero occlusion query images, $5$ images per category randomly drawn from the zero occlusion query images, are employed to synthesize the occluded query images.
In the second experiment, all of the zero occlusion query images are used to generate the occluded query images.
The examples of the occluded query images for the second experiment are shown in \figurename{~\ref{fig:massive_patchs_list}}.
The hyperparameter $\lambda$ for the proposed SDBE approaches is fixed to $0.005$ for both experiments.}

The results of the first experiment are tabulated in \tablename{~\ref{tab:ILSVRC_exp1}}.
It is evident that the proposed SDBE approaches significantly boost the classification accuracy, e.g., around $22.25\%$ increase in the classification accuracy for $20\%$ occlusion.
We can also observe that for zero occlusion, the SDBE\_L1 and SDBE\_L2 have very small drops ($0.47\%$ and $0.5\%$, respectively) compared to the original network.
This result again demonstrate that the proposed SDBE-based classification scheme is a unified scheme for both occluded and occlusion-free images.
In \tablename{~\ref{tab:ILSVRC_exp1}}, the SDBE\_L1 and SDBE\_L2 show almost the same classification accuracies.
This is because the original softmax classifier of the ResNet-152 network is a very powerful classifier, since it is trained on the large-scale training set of the ILSVRC 2012 dataset.

{The results of the second experiment are reported in \tablename{~\ref{tab:ILSVRC_exp2}}. It can be seen that
SDBE\_L2 achieves $21.8\%$ and $12.62\%$ performance increases at $20\%$ and $10\%$ occlusion, respectively.
The significant improvements demonstrate that the proposed SDBE\_L2
approach is able to deal with a large number of occlusion patterns on a large-scale dataset.}
\begin{table*}
	\renewcommand{\arraystretch}{1.4}
	\centering
	\fontsize{8}{8}\selectfont
	\begin{tabular}{l|l|l|l|l|l|l|l|l|l|l|l|l|l|l}
		\hline
		\multirow{2}{*}{OR}   & \multirow{2}{*}{method} & \multicolumn{12}{c|}{occlusion patch}                                                                                 & \multirow{2}{*}{Avg.} \\ \cline{3-14}
		&                         & 1       & 2       & 3       & 4       & 5       & 6       & 7       & 8       & 9       & 10      & 11      & 12      &                       \\ \hline
		\multirow{2}{*}{0\%}  & original                &\textemdash&\textemdash&\textemdash&\textemdash&\textemdash&\textemdash&\textemdash &\textemdash&\textemdash&\textemdash&\textemdash&\textemdash& 75.39                                    \\ \cline{2-15} 
		& SDBE\_L2                &\textemdash&\textemdash&\textemdash&\textemdash&\textemdash&\textemdash&\textemdash&\textemdash&\textemdash &\textemdash&\textemdash&\textemdash& 74.82                                    \\ \hline
		\multirow{2}{*}{10\%} & original                & 49.56 & 10.29 & 53.52 & 50.62 & 52.53 & 50.43 & 49.37 & 41.31 & 57.90 & 59.68 & 54.05 & 61.36 & 49.22                                    \\ \cline{2-15} 
		& SDBE\_L2                & 62.10 & 51.46 & 59.82 & 61.51 & 63.14 & 62.43 & 61.74 & 60.50 & 64.79 & 65.65 & 62.15 & 65.98 & 61.77                                    \\ \hline
		\multirow{2}{*}{20\%} & original                & 23.86 & 1.75  & 36.39 & 38.50 & 27.38 & 18.05 & 23.99 & 15.62 & 32.40 & 38.21 & 24.08 & 42.70 & 26.91                                    \\ \cline{2-15} 
		& SDBE\_L2                & 49.65 & 33.28 & 49.43 & 52.64 & 48.27 & 44.88 & 50.40 & 45.12 & 52.92 & 55.80 & 49.20 & 52.92 & 48.71                                    \\ \hline
	\end{tabular}
	\caption{Comparison of classification accuracy (\%) for the conventional scheme and the proposed SDBE\_L2 scheme on ILSVRC2012 dataset (OR: occlusion ratio). The pre-trained ResNet-152 network is adopted as the base CNN. The occlusion patches are shown in \figurename{\ref{fig:massive_patchs_list}(a)}}
	\label{tab:ILSVRC_exp2}
\end{table*}

We should also note that, in the above experiments, the classification accuracy of the original ResNet-152 network for the occlusion-free query images is slightly worse than the result reported in \cite{he2016deep}.
This may be caused by the inaccuracy in the re-implementation of the ResNet-152 network in MatConvNet \cite{vedaldi15matconvnet}.
Nevertheless, such a small difference does not devalue the merits of the proposed SDBE-based classification scheme.

\section{Comparison between SDBE\_L1 and SDBE\_L2} \label{sec:further_discussion}
We have introduced two SDBE approaches: SDBE\_L1 and SDBE\_L2.
In this section, to provide a useful guidance for practical applications, we will make a comparison between these two approaches from two aspects: classification accuracy and computaional complexity.

\subsection{Classification accuracy}
We consider two factors in the comparison of classification accuracy.
The first is the set of extra image pairs used to construct the OED.
For case 2 in Section \ref{exp:estmt_accrcy}, the SDBE\_L1 achieves better results than the SDBE\_L2.
The second is the classifier following the SDBE procedure.
For a weak classifier, such as the NN used in Section \ref{exp:estmt_accrcy} and the softmax used in Section \ref{exp:caltech101dataset}, the SDBE\_L1 achieves better classification results, while for a powerful classifier, such as the linear SVM used in Section \ref{exp:caltech101dataset} and the softmax trained on the ILSVRC2012 training set and used in Section \ref{exp:imagenet}, the SDBE\_L2 achieves almost the same (or slightly better in some test points) classification results as the SDBE\_L1.
This can be attributed to the powerfulness of the classifier.
The powerful classifier can generate the category boundaries with a large margin to the training DFVs such that small discrepancy between the SDBE\_L1 estimation and the SDBE\_L2 estimation will not lead to a noticeable difference in the classification results.

\subsection{Computational complexity}
Although, in many cases of the above experiments, the SDBE\_L2 a little bit underperforms the SDBE\_L1 in terms of classification accuracy, it has much lower computational complexity.
As mentioned in Section \ref{sec:regularization}, the SDBE\_L2 has a computational complexity of $\mathcal{O}(n)$ for the testing process.
For instance, in the first experiment of Section \ref{exp:imagenet}, the average execution time per query image for the SDBE\_L2 without GPU acceleration is around $0.005$s, as a comparison, the SDBE\_L1 costs around $10$s. 

In addition, the SDBE\_L2 can be implemented as a fully connected linear network layer.
Rewrite $\mathbf{P}$ as $\mathbf{P}=[\mathbf{P}_{\alpha}^T\; \mathbf{P}_{\beta}^T]^T$, where $\mathbf{P}_{\alpha}\in\mathbb{R}^{n_{\mathbf{A}}\times m}$ and $\mathbf{P}_{\beta}\in\mathbb{R}^{p_{\mathbf{B}}\times m}$.
Accordingly, equation (\ref{equ:l2norm_sltn}) becomes $\hat{\boldsymbol{\omega}}=[\hat{\boldsymbol{\alpha}}^T\; \hat{\boldsymbol{\beta}}^T]^T=[(\mathbf{P}_{\alpha}\mathbf{v}_i)^T\; (\mathbf{P}_{\beta}\mathbf{v}_i)^T]^T$.
Then, according to equation (\ref{equ:dfv_rcvry}), $\mathbf{v}_{0i}$ can be estimated by
\begin{equation}\label{eqn:v0i_networklayer}
\hat{\mathbf{v}}_{0i}=\mathbf{A}\mathbf{P}_{\alpha}\mathbf{v}_i.
\end{equation} 
Actually, this equation can be implemented as a fully connected layer with connection weight matrix $\mathbf{W}=\mathbf{A}\mathbf{P}_{\alpha}$.
{The weight maxtrix $\mathbf{W}$ can be computed in advance during the training process, and thus, the processing time and memory consumption of the SDBE\_L2 for each query image are independent of the sizes of the CD and OED.
By using the linear network layer implementation and accelerating with a GeForce GTX 1080 Ti GPU, the average execution time per query image for the SDBE\_L2 procedure is less than $0.002$s for the second experiment in Section \ref{exp:imagenet}, on which $\mathbf{W}$ has the largest number of rows ($18000$) in all the experiments.
Another advantage of this implementation is that the computational complexity of the testing process is independent of the size of the OED.}

Low computational complexity is an attribute preferred in the classification of large-scale datasets. 
Therefore, for large-scale datasets, where sufficient training images are available to train a powerful classifier, the SDBE\_L2 is a better choice.
On the contrary, for the classification of small-scale datasets or a task insensitive to execution time, the SDBE\_1 can be selected to achieve better classification results.

\section{Conclusion}\label{sec:Conclusion}
In this paper, we have proposed an SDBE-based classification scheme.
The proposed SDBE-based classification scheme has the following characteristics.
\begin{enumerate}
	\item It is a unified scheme for the classification of both occluded and occlusion-free images on any image datasets and can significantly boost the classification performance of the occluded images.
	\item Dealing with the occlusion in the deep feature space requires a small number of occluded images for training. 
	A dozen of occluded images for each occlusion pattern are able to improve the classification accuracy significantly.
	\item It requires less effort to accommodate new occlusion patterns.
	To adapt to a new occlusion pattern, it does not require to re-train or fine-tune the base CNN and only needs to insert the OEVs associated with the new occlusion pattern into the OED.
	Re-training or fine-tuning the base CNN is not only time-consuming but also expensive in some applications since, for instance, delivery of the updated CNN model to client ends is expensive, and sometimes unacceptable, in cloud based applications.
\end{enumerate}
{Although the above experiments were conducted with all the occlusion patterns in the query images available to the OED, the proposed SDBE approaches can consistently improve the classification accuracy for the occlusion patterns unseen to the OED (see Section S.II in the supporting document for additional experiments on such a situation).
This means that the proposed SDBE approach can work with any occluded images and it is able to be plugged into any well-trained CNNs.
However, lots of effort is still required towards real applications, such as the optimization of the CD and OED, enhancement of the performance for the occlusion patterns unavailable to the OED (see Section S.II in the supporting document for some clues), and the extension of the proposed approach to other tasks, such as object detection.}

{
\bibliographystyle{IEEEtran}
\bibliography{cyb_rcg_occ}
}

	

\vfill
\end{document}